%% file: main.tex
\documentclass{article}

\usepackage[preprint]{neurips_2026}

\makeatletter
\def\@noticestring{}
\makeatother

\input{math_commands}

\usepackage[utf8]{inputenc} 
\usepackage[T1]{fontenc}    
\usepackage{url}            
\usepackage{booktabs}       
\usepackage{amsfonts}       
\usepackage{nicefrac}       
\usepackage{microtype}      
\usepackage{svg}
\usepackage{float}
\usepackage{algorithm}
\usepackage{algpseudocode}
\usepackage[dvipsnames]{xcolor}
\hypersetup{
colorlinks=true,
linkcolor=magenta,
citecolor=magenta,
}

\newcommand{\ModelName}{EP-Flow\xspace}

\title{EP-Flow: Disordered Crystal Structure Prediction \\without Site-Level Annotations}

\author{Qiuliang Liu$^{1, 2, \dagger}$, Liming Wu$^{3, \dagger}$, Qi Li$^{1, 2}$, Zhonglong Peng$^{1, 2}$, Chang Chen$^{1, 2}$ \\ 
\textbf{Xiaolong Chen$^{1, 2}$, Wenbing Huang$^{3, *}$, Shifeng Jin$^{1, 2, *}$}  \vspace{0.2cm}\\
$^1$ Institute of Physics, Chinese Academy of Sciences, Beijing, China \\
$^2$ University of Chinese Academy of Sciences, Beijing, China \\
$^3$ Gaoling School of Artificial Intelligence, Renmin University of China \vspace{0.2cm} \\
$^\dagger$ These authors contributed equally to this work. \\
$^*$ Corresponding authors.}

\begin{document}

\maketitle

\begin{abstract}\label{sec:abstract}

Generative models have made rapid progress in ordered crystal structure prediction, yet many functional materials are intrinsically disordered, with substitutional mixing, vacancies, or interstitial species controlling their properties. Existing crystal generators either assume deterministic site occupations or require site-level disorder annotations, which are often unavailable when the chemical formula is the primary input. We formulate disordered crystal structure prediction through an Occupancy Distribution Matrix (ODM), a continuous site-by-species representation that unifies ordered crystals, solid solutions, vacancy disorder, and interstitial occupancy. A valid ODM must satisfy coupled site-wise occupancy, mass-conservation, and non-negativity constraints, placing each sample on a formula-dependent transportation polytope. We propose Entropic Polytope Flow (EP-Flow), a marginal-constrained flow matching framework that canonicalizes heterogeneous polytopes into a shared double-centered space, learns a marginal-preserving flow, and recovers feasible occupancies through a Sinkhorn inverse map. By jointly generating occupancies, fractional coordinates, and lattice parameters, EP-Flow achieves state-of-the-art performance on formula-conditioned disordered CSP benchmarks derived from COD and MPDS, substantially outperforming adapted ordered-crystal generators. Analyses further show that EP-Flow recovers sparse and chemically meaningful local disorder patterns rather than merely matching global composition statistics.
\end{abstract}

\section{Introduction}\label{sec:intro}

Crystalline materials underpin technologies ranging from semiconductors and batteries to solid electrolytes, catalysts, magnets, and superconductors. Recent machine learning advances have accelerated Crystal Structure Prediction (CSP) and generative materials design, with models such as CDVAE~\cite{xie2022cdvae}, GNoME~\cite{merchant2023scaling}, DiffCSP~\cite{jiao2023crystal}, FlowMM~\cite{miller2024flowmm} and MatterGen~\cite{zeni2025mattergen}
demonstrating remarkable progress. However, most existing methods are built around ordered structures, often derived from DFT calculations at 0 K, where each crystallographic site is occupied by a single deterministic element. Real materials frequently exhibit crystallographic disorder at synthesis or operating conditions, including substitutional mixing, vacancies, and interstitial species. This mismatch is substantial rather than marginal: nearly 47\% of verified ICSD structures~\cite{belsky2002icsd} contain some form of disorder. Such disorder can govern ionic transport, thermal stability, magnetism, and electronic behavior, motivating generative models that explicitly handle disordered crystals.

Disordered CSP is not merely ordered CSP with soft labels. In conventional ordered CSP, once a cell size or formula-unit convention is specified, the target composition fixes the species to be placed, and each generated site is assigned one deterministic species. In disordered crystals, the model must also infer how stoichiometric mass is allocated over crystallographic sites: multiple species may share a site, a site may be partially vacant, or an interstitial position may be fractionally occupied.
For common substitutional solid solutions, the site budget is usually implied by the formula. However, for vacancy and interstitial disorder, the reported composition is insufficient to specify all crystallographic sites. To resolve this, we use an extended stoichiometric vector $\vc$ by incorporating a vacancy pseudo-species ($\mathrm{Vac}$). This ensures every site is fully accounted for, satisfying $\sum_{j=1}^{M} c_j = N$, where $N$ is the total site count, $M$ is the number of candidate species (including $\mathrm{Vac}$), and $c_j$ is the target count for species $j$. This yields an Occupancy Distribution Matrix (ODM) encoding site-wise distributions. A valid ODM must satisfy three coupled constraints: site-wise full occupancy, mass conservation matching $\mathbf{c}$, and non-negativity.
We illustrate diverse types of disordered structures in \cref{app:details_disorder}, and detail the mapping between macroscopic formulas and microscopic site counts in \cref{app:details_for_site}.

\begin{wrapfigure}{r}{0.43\textwidth}
\vspace{-0.5cm}
\begin{center}
\includegraphics[width=0.43\textwidth]{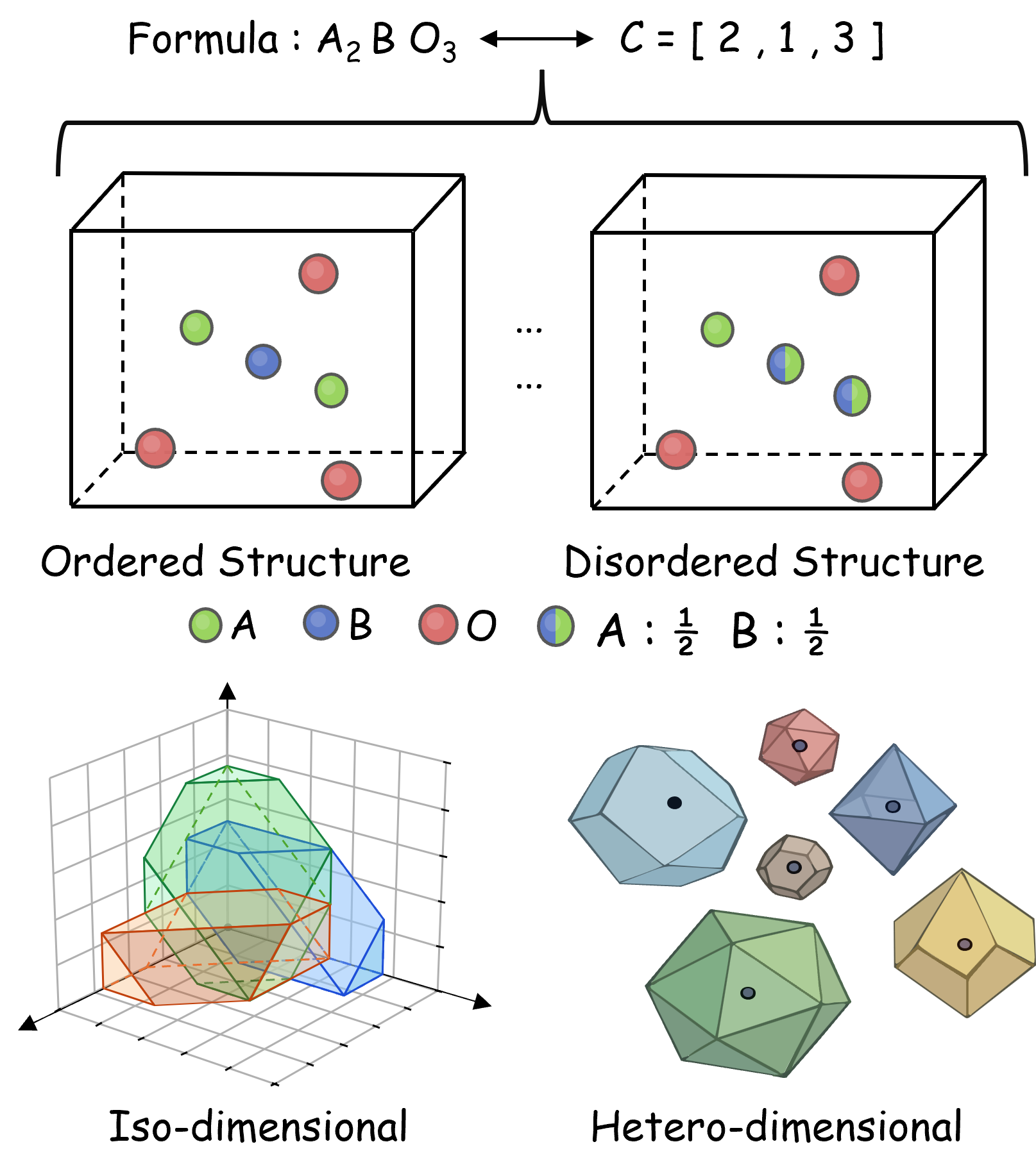}
\end{center}
\vspace{-0.5cm}
\caption{Formula transportation polytopes.}
\label{fig:transpotatioon polytopes}
\vspace{-0.35cm}
\end{wrapfigure}

Together, these requirements define a formula-dependent transportation polytope, as illustrated in~\cref{fig:transpotatioon polytopes}. Unlike standard crystal generators that predict site labels independently or correct compositions after generation, ODM generation is globally coupled: changing one site occupancy changes the feasible allocations at other sites. Moreover, the feasible domain changes with both composition and site budget. Standard diffusion or flow models are therefore difficult to apply directly, because their trajectories must respect a heterogeneous family of constrained domains rather than a single Euclidean space.

To address this challenge, we propose EP-Flow, a marginal-constrained flow matching framework for disordered CSP without site annotation. EP-Flow canonicalizes each formula-dependent transportation polytope into a shared double-centered linear subspace by applying log-domain double centering to the ODM. In this space, the model learns a marginal-preserving flow whose velocity field remains in the zero-marginal tangent space. A Sinkhorn-based inverse map then recovers an ODM on the target transportation polytope, enforcing row-sum, column-sum, and non-negativity constraints. Built on this occupancy formulation, EP-Flow jointly generates occupancies, fractional coordinates, and lattice parameters.

We evaluate EP-Flow on large-scale COD~\cite{gravzulis2012cod} and MPDS~\cite{MPDS} benchmarks covering substitutional disorder, vacancy-inclusive disorder, and systems with up to N=50 crystallographic sites. Compared to the adapted baselines, EP-Flow achieves state-of-the-art performance under this no-site-annotation CSP setting, substantially improving structure match rate, occupancy accuracy, formula consistency, and validity. Case studies further show that EP-Flow recovers sparse and chemically meaningful disorder motifs rather than merely matching global composition statistics.

In summary, our contributions are threefold: \textbf{(1)} we formulate disordered CSP without site annotation as generation over formula-dependent transportation polytopes using an ODM representation that unifies ordered crystals, substitutional disorder, vacancy disorder, and interstitial occupancy; \textbf{(2)} we introduce EP-Flow, a marginal-constrained flow matching framework with double-centered canonicalization and Sinkhorn recovery; and \textbf{(3)} we construct COD and MPDS benchmarks with occupancy-sensitive metrics, demonstrating state-of-the-art performance for formula-conditioned disordered CSP without site-level disorder priors.

\section{Related Work}
\textbf{Ordered Crystal Structure Prediction.} Generative modeling for crystals has progressed from latent-variable models~\cite{xie2022cdvae,luo2024cond_cdvae} to diffusion models~\cite{zeni2025mattergen,jiao2023crystal,lin2024equicsp,wu2025dao,joshi2025all,yi2026crystaldit} and, more recently, flow-based and Bayesian flow approaches~\cite{luo2024crystalflow,miller2024flowmm,li2025powder,jiao2025mofbfn,ruple2025symmbfn,wu2025crysbfn}. These methods have substantially improved the modeling of periodic atomic coordinates, but they are fundamentally designed for \emph{ordered} crystals, where each site is assigned a single discrete species. By contrast, \ModelName targets disordered crystal generation, where occupancies are fractional and coupled by global stoichiometric constraints.

\textbf{Disordered Crystal Structure Prediction.}
Recognizing the limitations of ordered models, recent work has started to address disordered structures explicitly. Dis-Gen~\cite{petersen2025dis_gen} represents crystals in a tabular form and reconstructs them with a VAE-style architecture, while DMFlow~\cite{wu2026dmflow} uses flow matching to model substitutional and positional disorder under simplex constraints. However, these methods still rely on site-level disorder annotations, e.g., which sites are doped and what occupancies they should carry. This assumption breaks down in the more challenging setting considered here, where only the global chemical formula is given. In contrast, \ModelName models the full occupancy distribution jointly, so that local full-occupancy and global mass-conservation constraints are handled within a unified generative process without requiring explicit site-specific supervision.

\textbf{Constrained Generative Modeling on Manifolds.} More broadly, our work is related to constrained generative modeling beyond materials science. Prior studies have extended diffusion and flow-based models to Riemannian manifolds, simplices, and more general polytopes~\cite{de2022riemannian,stark2024dirichlet,gat2024discrete,davis2024fisher,diederen2025flows}, using projection-based corrections~\cite{utkarsh2025physics} or barrier/mirror-map formulations~\cite{fishman2023diffusion,liu2023mirror}. These strategies, however, can either break trajectory smoothness or become numerically fragile near sharp polytope boundaries. Our approach instead develops a marginal-preserving flow in a shared double-centered space, followed by a Sinkhorn inverse map, yielding exact transportation-polytope constraints.

\section{Preliminaries}\label{sec:prelimilary}
We briefly review the two ingredients underlying our method: transportation-polytope constraints and the double-centered zero-marginal space that provides a shared representation across formulas.

\subsection{Transportation Polytopes and Sinkhorn Projection}
Let $\vr\in\mathbb{R}^N_{+}$ and $\vc\in\mathbb{R}^M_{+}$ be prescribed marginal vectors with equal total mass, i.e., $\bm{1}_N^\top \vr=\bm{1}_M^\top \vc$. The associated transportation polytope is defined as the set of non-negative matrices satisfying these exact marginal constraints:
\begin{equation}
\mathcal{P}(\vr,\vc)
\;=\;
\Big\{\mP\in\mathbb{R}^{N\times M}_{+}\; \big|\;
\mP\bm{1}_M=\vr,\;
\bm{1}_N^\top\mP=\vc^\top
\Big\}.
\end{equation}

Given a non-negative matrix $\mX\in\mathbb{R}^{N\times M}_{+}$, the Sinkhorn algorithm~\cite{sinkhorn1964relationship} alternates row and column rescaling to project $\mX$ onto $\mathcal{P}(\vr,\vc)$. We denote the resulting differentiable operator by $\mathcal{S}(\mX;\vr,\vc)\in\mathcal{P}(\vr,\vc)$. In practice, a finite number of iterations is sufficient to satisfy the marginal constraints up to a prescribed tolerance~\cite{cuturi2013sinkhorn}. For strictly positive inputs, this projection is differentiable and coincides with the unique Kullback-Leibler (KL) projection onto the target polytope~\cite{peyre2019computational}.

\subsection{Double-Centering Operator and Zero-Marginal Space}
To map complex constrained distributions into a tractable representation, we rely on the double-centering operator. For any arbitrary matrix $\mX \in \mathbb{R}^{N \times M}$, the double-centering operator $\mathfrak{D}(\cdot)$ subtracts its row and column means, expressed as:
\begin{equation}
\mathfrak{D}(\mX) = (\mathbf{I}_N - \frac{1}{N}\mathbf{1}_N\mathbf{1}_N^\top)\mX(\mathbf{I}_M - \frac{1}{M}\mathbf{1}_M\mathbf{1}_M^\top).
\label{eq:double_center}
\end{equation}

The image of this operator defines the zero-marginal subspace:

\begin{equation}
    \mathcal{M}_0(N,M)
    \triangleq 
    \Big\{
    \mZ\in\mathbb{R}^{N\times M}
    \;\big|\; 
    \mZ\bm{1}_M=\bm{0}_N,\; 
    \bm{1}_N^\top\mZ=\bm{0}_M^\top 
    \Big\}.
\end{equation}
Because $\mathcal{M}_0(N,M)$ is a linear subspace, it serves as a well-behaved Euclidean manifold whose tangent space is isomorphic to itself. This makes $\mathcal{M}_0(N,M)$ a convenient Euclidean space for defining generative vector fields while preserving the zero-sum structure induced by the constraints.


\section{Proposed Approach: EP-Flow}\label{sec:method}
\subsection{Task Definition}

\begin{figure}[t!]
    \centering
    \includegraphics[width=0.99\textwidth]{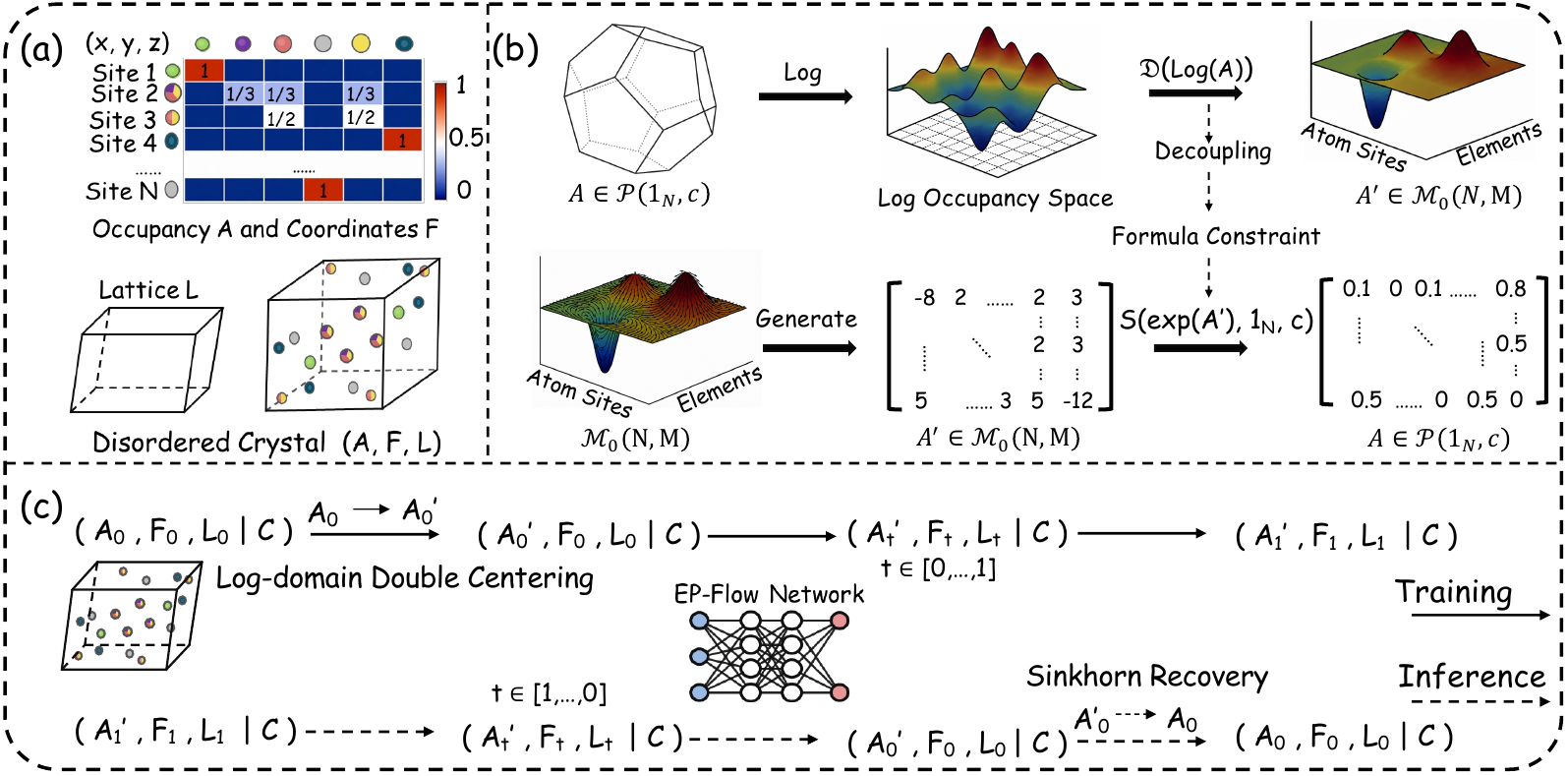}
    \vspace{-0.1cm}
    \caption{
    \textbf{Overview of EP-Flow for disordered crystal generation.}
    \textbf{(a)} Disordered crystal represented by $(\mA,\mF,\vl)$.
    \textbf{(b)} Occupancy matrix canonicalized from formula polytope $\mathcal{P}(\bm{1}_N,\vc)$ to double-centered space $\mathcal{M}_0(N,M)$ via log-domain double centering, which recovers back by Sinkhorn projection.
    \textbf{(c)} Joint flow matching over $(\mA',\mF,\vl)$ conditioned on $\vc$, with projection back to $\mA$ for the final structure.
    }
    \label{fig:odm_generation}
\end{figure}

We represent a disordered crystal as $\mathcal{D}=(\mA,\mF,\vl)$, where $\mA\in[0,1]^{N\times M}$ is the ODM over $N$ sites and $M$ species, $\mF\in[0,1)^{N\times 3}$ denotes fractional coordinates, and $\vl=(a,b,c,\alpha,\beta,\gamma)\in\mathbb{R}_+^6$ denotes the lattice parameters, as shown in \cref{fig:odm_generation}a. Given a chemical formula, we encode its stoichiometry as $\vc\in\mathbb{R}^M$, where $c_j$ is the total occupancy of species $j$ over all sites. Physical validity requires $\mA$ to satisfy row-sum constraints $\sum_{j=1}^M A_{ij}=1$ for full site occupancy, column-sum constraints $\sum_{i=1}^N A_{ij}=c_j$ for global mass conservation, and non-negativity $A_{ij}\ge 0$. Hence, $\mA$ lies in a formula-dependent transportation polytope $\mathcal{P}(\bm{1},\vc)$. By introducing vacancy as a pseudo-species, this representation naturally covers ordered crystals, substitutional disorder, and vacancy disorder.


Disordered CSP is then formulated as learning the conditional distribution $p(\mA,\mF,\vl\mid\vc)$ by flow matching (details see \cref{app:flow_matching}). This differs fundamentally from conventional CSP formulations that assume fixed site-wise assignments and model only $p(\mF,\vl\mid\mA)$. Here the model receives only $\vc$ and must generate a full crystal triplet whose occupancy matrix remains inside $\mathcal{P}(\bm{1},\vc)$ while coordinating consistently with atomic positions and lattice geometry. This shift from explicit site labels to a continuous, constraint-aware occupancy representation is the key to exploring a much richer space of disordered materials, without assuming prior knowledge of specific doping patterns or site preferences.

\subsection{Occupancy Distribution Matrix Generation}


The primary challenge in generating $\mA\in[0,1]^{N\times M}$ is that its support, the transportation polytope $\mathcal{P}(\bm{1},\vc)$, changes with the stoichiometry $\vc$ and the number of sites $N$. ODMs therefore live on a heterogeneous family of constrained domains with no shared geometry, making direct generative modeling difficult. We resolve this by canonicalizing each ODM into the shared double-centered space $\mathcal{M}_0(N,M)$, where all matrices satisfy zero-sum marginals regardless of the underlying formula.

\begin{definition}[Space Canonicalization]
Given an ODM $\mA\in \mathcal{P}(\bm{1},\vc)$, we define a forward mapping $\Psi:\mathcal{P}(\bm{1},\vc)\to \mathcal{M}_0(N,M)$ by
\begin{equation}
\label{eq:psi}
    \mA'=\Psi(\mA)\triangleq \mathfrak{D}(\log \mA)
\end{equation}
where $\mathfrak{D}$ is the double-centering operator, as shown in \cref{eq:double_center}. The image lies in $\mathcal{M}_0(N,M)$.


Intuitively, double-centering removes the additive row- and column-wise components of $\log\mA$, which can be interpreted as site- and species-specific chemical potentials, and retains only the residual interaction structure in a stoichiometry-invariant space~\cite{torgerson1952multidimensional}. In practice, we apply $\epsilon$-smoothing to prevent numerical instability during logarithm operations. Algorithm details are provided in \cref{alg:log_dc}.
\end{definition}

To recover a valid ODM under a target stoichiometry $\vc$, we map back from $\mathcal{M}_0(N,M)$ using a Sinkhorn projection~\cite{sinkhorn1964relationship}:
\begin{equation}
\label{eq:inv-psi}
    \mA=\Psi^{-1}(\mA')\triangleq \mathcal{S}\!\left(\exp(\mA'),\,\bm{1},\,\vc\right).
\end{equation}
In this context, $\mathcal{S}$ acts as a projection that finds the matrix in $\mathcal{P}(\bm{1}, \vc)$ that minimizes the Kullback-Leibler (KL) divergence to the exponentiated centered matrix $\exp(\mA')$, thereby restoring the physical stoichiometry. We illustrate the process in \cref{fig:odm_generation}b, and provide more details in \cref{alg:sinkhorn_recovery}.

As established in \cref{eq:psi,eq:inv-psi}, the function $\Psi$ is a bijective mapping, as stated in the following proposition:
\begin{proposition}[Bijectivity of Canonicalization]
For any fixed stoichiometry vector $\vc$, the mapping $\Psi$ establishes a one-to-one correspondence between the interior of the transportation polytope $\mathcal{P}(\bm{1}, \vc)$ and the double-centered space $\mathcal{M}_0(N,M)$. Specifically, for any $\mA' \in \mathcal{M}_0(N,M)$, there exists a unique matrix $\mA \in \text{int}(\mathcal{P}(\bm{1}, \vc))$ such that $\Psi(\mA) = \mA'$, which is uniquely recoverable via the Sinkhorn operator: $\mA = \mathcal{S}(\exp(\mA'), \bm{1}, \vc)$. This bijectivity ensures that the generative trajectory in $\mathcal{M}_0(N,M)$ maps to a unique and physically consistent path in the valid ODM space. A proof is provided in \cref{app:proof-bijection}.
\end{proposition}

With this shared representation, we learn the centered ODM distribution by flow matching. We use the linear path $\mA'_t=(1-t)\mA'_0+t\mA'_1$ for $t\in[0,1]$, where $\mA'_1=\Psi(\mA_1)$ comes from data and $\mA'_0$ is sampled from a prior supported on $\mathcal{M}_0(N,M)$. Since $\mathcal{M}_0(N,M)$ is linear, the entire path remains in $\mathcal{M}_0(N,M)$ as long as the predicted velocity is projected back by $\mathfrak{D}$, and we provide the proof of trajectory feasibility in ~\cref{app:proof-Trajectory}. We additionally apply a formula-derived mask to restrict training to site--species pairs that are feasible in the data. The resulting target velocity is the constant field $
u_t^{\mA'}=\mA'_1-\mA'_0$, 
which supervises the projected network output in $\mathcal{M}_0(N,M)$.

\subsection{Fractional Coordinates Generation}


Fractional coordinates $\mF\in[0,1)^{N\times 3}$ are periodic along each lattice direction, so we define their flow on the $N$-torus $(\mathbb{T}^3)^N$. Given a noise sample $\mF_0$ and a data sample $\mF_1$, the target velocity is the shortest periodic displacement
\begin{equation}
\begin{aligned}
    u_t^{\mF}
    =
    \Delta_{\mathbb{T}^3}(\mF_1,\mF_0)
    =
    \big((\mF_1-\mF_0+0.5)\bmod 1\big)-0.5 .
\end{aligned}
\end{equation}
This torus-aware displacement is the geodesic direction on the periodic coordinate space. The corresponding interpolation is
\begin{equation}
    \mF_t =
    \left(\mF_0 + t u_t^{\mF}\right) \bmod 1,
    \qquad t \in [0,1].
\end{equation}
The model is trained to predict $u_t^{\mF}$ from $\mF_t$, ensuring periodicity is respected throughout the trajectory.

\subsection{Lattice Parameters Generation}
Following FlowMM~\cite{miller2024flowmm}, we model the lattice parameters $\vl=(a,b,c,\alpha,\beta,\gamma)$ in an unconstrained latent space for stable learning. The lengths $(a,b,c)$ are parameterized in log space to guarantee positivity, while the angles $(\alpha,\beta,\gamma)$ are parameterized by unconstrained logits and mapped to the Niggli-reduced range $[60^\circ,120^\circ]$. This yields a canonical lattice representation (i.e., $\vl^{\mathrm{unc}}$) while preserving the coupled statistics between cell geometry and atomic arrangement. Flow matching supervision is defined in the unconstrained space with target velocity $u_t^{\vl}=\vl_1^{\mathrm{unc}}-\vl_0^{\mathrm{unc}}$.

\subsection{Unified Training Objective}
The model is trained by jointly regressing the target velocity fields of all structural components. At time $t\in[0,1]$, the intermediate state is $\mathcal{D}_t=(\mA_t,\mF_t,\vl_t)$, and the network $v_\theta(\mathcal{D}_t,t\mid\vc)$ outputs three velocity heads for occupancy, coordinates, and lattice variables, conditioned on the stoichiometry $\vc$. 
To account for the different scales of lattice lengths and angles, the lattice head is further split into length and angle components, denoted by $v_\theta^{\mathrm{len}}$ and $v_\theta^{\mathrm{ang}}$. The overall objective is a weighted sum of mean-squared errors:
\begin{equation}
\begin{aligned}
    \mathcal{L}
    &= \lambda_{\mA'}\,\mathbb{E}_{t,\, \mathcal{D}_1,\, \mathcal{D}_t}
    \Big[\big\| v_\theta^{\mA'}(\mathcal{D}_t, t \mid \vc) - u_t^{\mA'} \big\|_2^2\Big] + \lambda_{\mF}\,\mathbb{E}_{t,\, \mathcal{D}_1,\, \mathcal{D}_t}
    \Big[\big\| v_\theta^\mF(\mathcal{D}_t, t \mid \vc) - u_t^\mF \big\|_2^2\Big] \\
    &+ \mathbb{E}_{t,\, \mathcal{D}_1,\, \mathcal{D}_t}
    \Big[\lambda_{\mathrm{len}}\big\| v_\theta^{\mathrm{len}}(\mathcal{D}_t, t \mid \vc) - u_t^{\mathrm{len}} \big\|_2^2 + \lambda_{\mathrm{ang}}\big\| v_\theta^{\mathrm{ang}}(\mathcal{D}_t, t \mid \vc) - u_t^{\mathrm{ang}} \big\|_2^2 \Big],
\end{aligned}
\end{equation}
where $\lambda_{\mA'},\lambda_\mF,\lambda_{\mathrm{len}},\lambda_{\mathrm{ang}}$ balance occupancy, coordinate, and lattice learning. Optimizing this objective encourages a physically consistent \emph{joint} distribution over these structural components under the stoichiometric constraint. The training and inference process is illustrated in \cref{fig:odm_generation}c.

\subsection{Stoichiometry-Conditioned Joint Velocity Network}
To predict the coupled velocity fields for occupancy, coordinates, and lattice parameters, we use a lightweight stoichiometry-conditioned message-passing network. At time $t$, the input consists of the centered occupancy state $\mA'_t$, fractional coordinates $\mF_t$, lattice state $\vl_t$, and stoichiometry vector $\vc$. We provide physical intuition and conceptual inspirations in \cref{app:physical_interpretations}. Specifically, we draw a heuristic analogy between statistical thermodynamics and our mathematical formulations, such as the logarithmic mapping, double-centering, and Sinkhorn recovery. 

For each site $i$, we construct an initial node embedding by combining a local occupancy embedding, a global formula embedding, a global lattice embedding, and a time embedding:
\begin{equation}
\vh_i^{(0)} = \phi_{\mathrm{init}}\!\left(\phi_{\mathrm{energy\_diff}}(\mA'_{t,i}),\,\phi_{\mathrm{form}}(\vc),\,\phi_{\mathrm{lat}}(\vl_t),\,\phi_{\mathrm{time}}(t)\right),
\end{equation}
where $\phi_*$ are lightweight MLPs. Within each crystal, we build a fully connected graph and perform several rounds of message passing. For an edge $(i,j)$, the message depends on the current node states, the coordinate displacement $\Delta \vf_{ij}$, the occupancy difference $\mA'_{t,i}-\mA'_{t,j}$, and the global formula/lattice context. Denoting the hidden state at layer $\ell$ by $\vh_i^{(\ell)}$, the updates are
\begin{align}
\vm_{ij}^{(\ell)} &= \phi_e\!\left(\vh_i^{(\ell)},\vh_j^{(\ell)},\mathrm{SinEmb}(\Delta \vf_{ij}),\phi_\text{energy\_diff}(\mA'_{t,i}-\mA'_{t,j}),\vc,\vl_t\right), \\
\vh_i^{(\ell+1)} &= \vh_i^{(\ell)} + \phi_v\!\left(\vh_i^{(\ell)},\operatorname{mean}_{j}\,\vm_{ij}^{(\ell)}\right),
\end{align}
where $\phi_e$, $\phi_\text{energy\_diff}$, and $\phi_v$ are MLPs, and $\mathrm{SinEmb}$ is the sinusoidal embedding used to encode periodic displacements~\cite{jiao2023crystal}. The residual update preserves local information while enabling each site to aggregate composition- and geometry-aware messages from the full crystal.

After the final message-passing layer, the backbone outputs three heads. Two site-wise heads predict the coordinate velocity $v_\theta^{\mF}$ and occupancy velocity $v_\theta^{\mA'}$, with the latter projected by $\mathfrak{D}$ back into $\mathcal{M}_0(N,M)$. A graph-level pooled representation is then used to predict the lattice velocity $v_\theta^{\vl}$. This network satisfies SO(3) invariance and translation invariance, as shown in \cref{app:proof-so3,app:proof-translation}.

\section{Experiments}\label{sec:experiments}

\subsection{Experimental Setup}
Currently, there are no publicly available benchmarks or models specifically designed for disordered occupancy distribution tasks. Therefore, we establish a comprehensive evaluation framework, encompassing self-constructed datasets, modified strong baselines, and a multi-tiered metric suite.

\textbf{Datasets.} 
Our benchmark is curated from two complementary databases, COD~\cite{gravzulis2012cod} and MPDS~\cite{MPDS}. The benchmark contains 17,239 structures from COD and 110,966 structures from MPDS. We organize four subsets: Sub-20 and Sub-50 for \textbf{substitutional disorder}, and Disorder-20 and Disorder-50 for the more general setting that additionally includes \textbf{vacancies}. Here, 20 and 50 indicate that the crystal contains at most 20 or 50 atomic sites, respectively. This split allows us to separately evaluate the disorder type and scaling with the number of sites. For each subset, we use an approximate 8:1:1 train/validation/test split. Detailed statistics and processing details are deferred to \cref{sec:datasets_supp}.

\textbf{Baselines.} Existing crystal generators are primarily designed for \emph{ordered} structures and therefore cannot be directly applied to our more challenging disordered setting. To construct strong and fair baselines, we adapt representative SOTA methods, including FlowMM~\cite{miller2024flowmm} and OMatG~\cite{hollmer2025omatg}, to operate on continuous site-wise occupancies rather than discrete one-hot species representations. On the generation side, these baselines first predict an unconstrained ODM in Euclidean space, which is then mapped to the positive orthant through a sigmoid transformation, enabling the subsequent application of the Sinkhorn projection. Furthermore, we adapt DMFlow~\cite{wu2026dmflow}, a strong baseline specifically designed for disordered crystal generation. We adopt the spherical reparameterization mechanism from its DNG, enabling the occupancy to diffuse on the sphere, which naturally satisfies the site-wise row-sum constraint. For all baselines, we apply the same formula-derived mask based on the ground-truth chemical composition as used in EP-Flow during both training and generation. In the inference phase, we uniformly apply the same Sinkhorn projection to project the occupancies onto the transportation polytope space, enforcing both the global formula sum and site-wise sum constraints. Further implementation details of all baseline adaptations are provided in ~\cref{app:baseline_adaptations}.

\textbf{Evaluation Metrics.} 
Our models are evaluated by four metrics: \textbf{(1)} Match Rate, which measures structure-level matching under the extended matcher; \textbf{(2)} RMSD, which quantifies coordinate accuracy for matched structures; \textbf{(3)} OccMAE, which measures the error in site-wise occupancy distributions; and \textbf{(4)} Validity, which measures the fraction of physically plausible outputs. The exact metric definitions, matcher tolerances, and implementation details are provided in \cref{sec:metric_supp}.

\begin{table}[t!]
\centering
\small
\caption{Disordered CSP Performance on the COD dataset. When match rate is zero (i.e., all failed), RMSD and OccMAE are undefined (N/A). The best results are marked in bold. ``Ada'' denotes the adapted baselines.}
\label{tab:COD_results}
\resizebox{\textwidth}{!}{
\begin{tabular}{llccccc}
\toprule
\textbf{Dataset} & \textbf{Model} & \textbf{Match Rate ($\uparrow$)} & \textbf{RMSD ($\downarrow$)} & \textbf{OccMAE ($\downarrow$)} & \textbf{Validity ($\uparrow$)} \\
\midrule
\multirow{4}{*}{\textbf{Sub-20}} & Ada-FlowMM &0.65\%	&\textbf{0.0146}	&\textbf{0.0002}	&87.40\% \\
& Ada-OMatG &10.21\%	&0.0298	&0.0147 &90.40\% \\
& Ada-DMFlow &21.61\%	&0.0580 &0.0189 &90.60\% \\
& EP-Flow & \textbf{45.39\%} & 0.0393 & 0.0111 & \textbf{95.10\%} \\
\midrule
\multirow{3}{*}{\textbf{Sub-50}} & Ada-FlowMM &0.67\% &0.0728 &\textbf{0.0092} &86.90\% \\
& Ada-OMatG &3.28\% &0.0459 &0.0218 &90.90\% \\
& Ada-DMFlow &10.84\%	&\textbf{0.0229}	&0.0503	&80.00\% \\
& EP-Flow & \textbf{26.77\%} & 0.0357 & 0.0107 & \textbf{92.60\%} \\
\midrule
\multirow{4}{*}{\textbf{Disorder-20}} & Ada-FlowMM &0.49\% &0.0365 &0.0407 &83.80\% \\
& Ada-OMatG &9.78\% &\textbf{0.0193} &\textbf{0.0154} &88.40\% \\
& Ada-DMFlow &15.16\% &0.0279 &0.0298 &87.20\% \\
& EP-Flow & \textbf{31.80\%} & 0.036 & 0.0163 & \textbf{89.20\%} \\
\midrule
\multirow{4}{*}{\textbf{Disorder-50}} & Ada-FlowMM &0.64\% &\textbf{0.0146} &\textbf{0.0079} &75.70\% \\
& Ada-OMatG &3.59\% &0.0472 &0.0154 &\textbf{79.70\%} \\
& Ada-DMFlow &9.74\% &0.0348 &0.0517 &74.10\% \\
& EP-Flow & \textbf{17.10\%} & 0.0334 & 0.0148 & 77.70\%  \\
\bottomrule
\end{tabular}}
\\[1pt]
\end{table}

\begin{table}[t!]
\centering
\caption{Disordered CSP Performance on the MPDS dataset. When match rate is zero (i.e., all failed), RMSD and OccMAE are undefined (N/A). The best results are marked in bold. ``Ada'' denotes the adapted baselines.}
\small
\label{tab:mpds_results}
\resizebox{\textwidth}{!}{
\begin{tabular}{llccccc}
\toprule
\textbf{Dataset} & \textbf{Model} & \textbf{Match Rate ($\uparrow$)} & \textbf{RMSD ($\downarrow$)} & \textbf{OccMAE ($\downarrow$)} & \textbf{Validity ($\uparrow$)} \\
\midrule
\multirow{4}{*}{\textbf{Sub-20}} & Ada-FlowMM &0.49\% &0.0335 &0.0457 &96.90\% \\
& Ada-OMatG &11.78\% &\textbf{0.0331} &0.0813 &88.30\% \\
& Ada-DMFlow &45.17\% &0.0416 &0.0096 &92.10\% \\
& EP-Flow & \textbf{60.82\%} & 0.0483 & \textbf{0.0075} & \textbf{97.80\%} \\
\midrule
\multirow{4}{*}{\textbf{Sub-50}} & Ada-FlowMM &0.18\% &0.0501 &0.0391 &\textbf{98.10\%} \\
& Ada-OMatG &5.25\% &\textbf{0.0242} &0.0594 &93.50\% \\
& Ada-DMFlow &25.00\% &0.0354 &0.0320 &81.50\% \\
& EP-Flow & \textbf{39.33\%} & 0.0538 & \textbf{0.0076} & 91.90\% \\
\midrule
\multirow{4}{*}{\textbf{Disorder-20}} & Ada-FlowMM &0.24\% &\textbf{0.0136} &0.0474 &\textbf{95.90\%} \\
& Ada-OMatG &9.91\% &0.0263 &0.0860 &92.40\% \\
& Ada-DMFlow &42.59\% &0.0329 &0.0211 &94.10\% \\
& EP-Flow & \textbf{57.74\%} & 0.0413 & \textbf{0.0092} & 94.90\% \\
\midrule
\multirow{4}{*}{\textbf{Disorder-50}} & Ada-FlowMM &0.13\% &0.0354 &0.0401 &\textbf{97.30\%} \\
& Ada-OMatG &3.76\% &\textbf{0.0178} &0.0460 &87.10\% \\
& Ada-DMFlow &23.84\% &0.0485 &0.0250 &84.10\% \\
& EP-Flow & \textbf{32.60\%} & 0.0550 & \textbf{0.0069} & 89.60\% \\
\bottomrule
\end{tabular}}
\\[1pt]
\vspace{-0.3cm}
\end{table}

\subsection{Main Results on Disordered Crystal Generation}

We summarize the quantitative performance in \cref{tab:COD_results,tab:mpds_results}. Three observations stand out: (1) \textbf{EP-Flow achieves the highest Match Rate across all evaluated subsets.} Across all eight subsets, EP-Flow reaches 17.10\%--60.82\% Match Rate, while FlowMM remains below 1\%, OMatG reaches 3.28\%--11.78\%, and DMFlow reaches 9.74\%--45.17\%. EP-Flow outperforms the strongest baseline, DMFlow, by 7.36--23.78 percentage points, suggesting the benefit of modeling dynamics on a formula-dependent transportation polytope rather than in a generic Euclidean space. This comparison should also be read together with RMSD and OccMAE: since both are averaged only over matched cases, a model with higher Match Rate can include harder successful matches and therefore exhibit slightly larger average RMSD, so these metrics should be interpreted jointly rather than in isolation. (2) \textbf{High validity does not imply correct structure recovery.} The adapted baselines can obtain high validity despite low Match Rate. For instance, on MPDS Disorder-20, FlowMM achieves 95.90\% validity but only 0.24\% Match Rate. This indicates that generating physically plausible structures alone does not ensure correct recovery of coupled site-wise occupancy and geometry. This further highlights the necessity of employing stringent structure-level matching metrics to evaluate the true predictive capacity of models in disordered CSP. (3) \textbf{Larger and more disordered systems are intrinsically harder.} From 20 to 50 sites, and from substitutional to vacancy-inclusive disorder subsets, both EP-Flow and the strong baseline Ada-DMFlow degrade consistently on both COD and MPDS, reflecting the greater combinatorial ambiguity of site assignment and the higher sparsity of occupancy patterns. Meanwhile, they perform consistently better on MPDS than on COD, suggesting that larger disordered datasets provide richer supervision for learning transferable site--species compatibility.

\subsection{Ablation Studies}

  
\begin{table}[t!]
\centering
\caption{Ablation results on MPDS-Sub-20.}
\label{tab:ablation}
\resizebox{0.98\textwidth}{!}{
\begin{tabular}{lccccc}
\toprule
\textbf{Variant} & \textbf{Match Rate ($\uparrow$)} & \textbf{RMSD ($\downarrow$)} & \textbf{OccMAE ($\downarrow$)} & \textbf{Validity ($\uparrow$)} \\
\midrule
w/o Formula & 20.01\% & 0.0529 & 0.0106 & 96.7\% \\
w/o Energy-diff & 35.23\% & 0.0527 & 0.0159 & 93.50\% \\
w/o Double Centering & 46.06\% & 0.0537 & 0.0115 & 91.0\% \\
w/o Projection & 15.78\% & \textbf{0.0434} & 0.1277 & 90.5\%\\
w/o DC \& Sinkhorn & 0.00\% & N/A & N/A & 0.00\% \\
\midrule
Full Model (Reference) & \textbf{60.82\%} & 0.0483 & \textbf{0.0075} & \textbf{97.8\%} \\
\bottomrule
\vspace{-0.6cm}
\end{tabular}}
\end{table}

To isolate the contribution of each core component, we conduct five ablations on MPDS-Sub-20 in \cref{tab:ablation}: (1) \textbf{w/o Formula}, which removes chemical-formula conditioning embedded at the network layers and sharply reduces Match Rate from 60.82\% to 20.01\%, showing that accurate occupancy prediction requires explicit stoichiometric guidance beyond local geometry; (2) \textbf{w/o Energy-diff}, which removes the relative energy-difference edge feature and lowers Match Rate to 35.23\% while increasing OccMAE to 0.0159, indicating that pairwise relative affinities provide important local signals for resolving fine-grained disorder. A more detailed physical interpretation is deferred to \cref{app:Energy-Differenc}; (3) \textbf{w/o Double Centering}, which removes the $\mathfrak{D}$ operator and learns in the log space, reducing Match Rate to 46.06\% and confirming that mapping formula-dependent polytopes into a shared centered space improves structural alignment; (4) \textbf{w/o Projection}, which learns directly in the raw bounded polytope and causes OccMAE to surge from 0.0075 to 0.1277, showing that the unflattened constraint geometry is too irregular for occupancy learning, even if the matched-sample RMSD appears competitive; and (5) \textbf{w/o DC \& Sinkhorn}, which removes both normalization and constraint projection and collapses completely to 0.00\% Match Rate, demonstrating that geometric normalization and marginal preservation are critically important for staying on the feasible manifold.

\subsection{More Analysis}

\begin{figure}[t!]
    \centering
    \vspace{0.0cm}
    \includegraphics[width=0.95\linewidth]{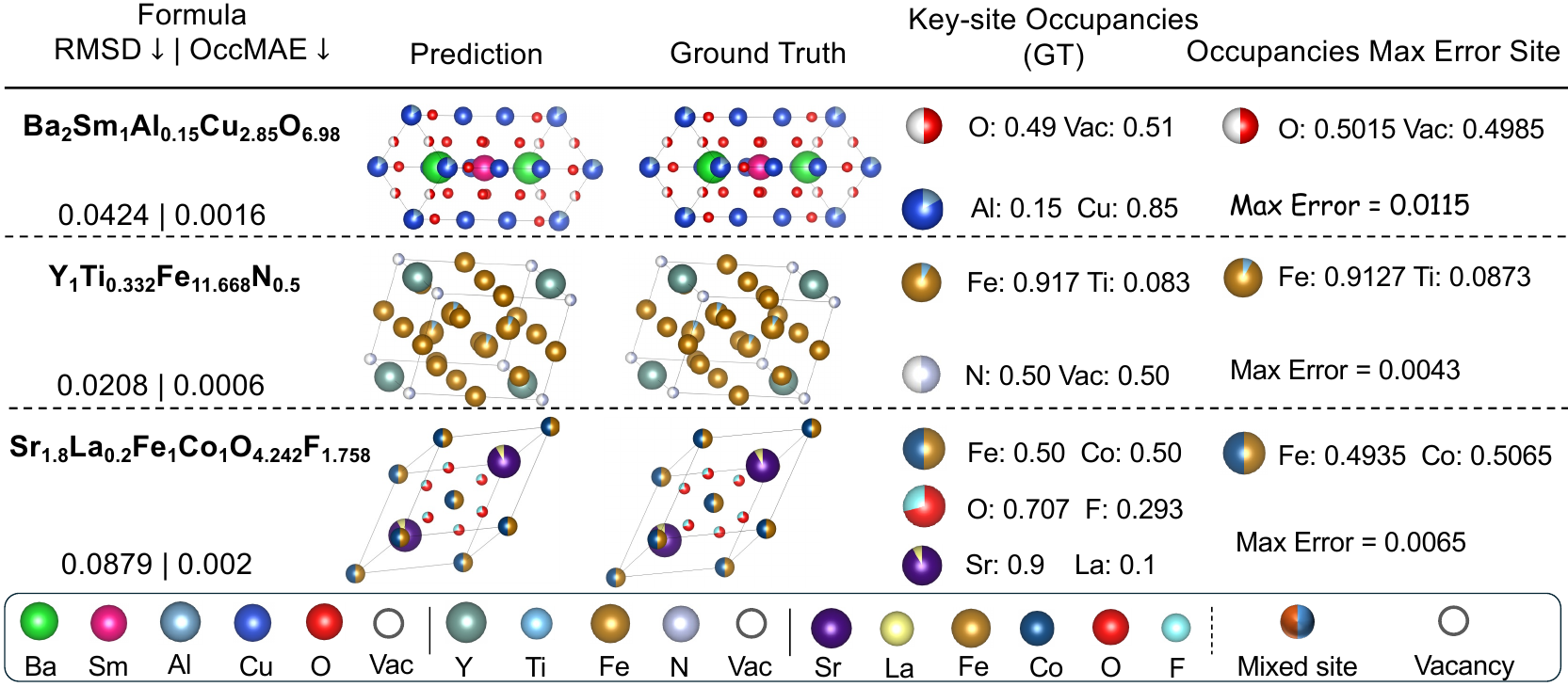}
    \vspace{-0.2cm}
    \caption{\textbf{Reconstruction results for disordered functional materials.} Each row compares predicted vs. ground-truth structures, with RMSD and OccMAE. Key-site occupancies are annotated; the rightmost column shows the site with largest occupancy error. Three disorder types are covered: oxygen-vacancy, interstitial/partial-occupancy, and mixed-cation/anion substitutional.}
    \vspace{-0.4cm}
    \label{fig:reconstruction_cases}
\end{figure}

\subsubsection{Case Studies on Functional Disordered Materials}

To assess whether EP-Flow captures realistic disorder beyond simple cation mixing, we examine three representative functional materials spanning substitutional, vacancy, and interstitial disorder. As shown in \cref{fig:reconstruction_cases}, these examples probe whether the model recovers not only average occupancies but also property-relevant local configurations.


\textbf{Layered cuprate with oxygen-vacancy disorder.} Ba$_{2}$SmAl$_{0.15}$Cu$_{2.85}$O$_{6.98}$ is an Al-substituted SmBa$_2$Cu$_3$O$_{7-\delta}$-type cuprate with dilute Cu/Al substitution and oxygen-vacancy disorder. In these high-$T_c$ materials, the functional impact of vacancies depends critically on whether they appear in the charge-regulating Cu--O chains or in the superconducting CuO$_2$ planes. EP-Flow localizes the vacancies outside the CuO$_2$ planes rather than distributing them randomly, indicating that it captures chemically meaningful defect preferences rather than only average geometry.


\textbf{ThMn$_{12}$-type magnet with interstitial disorder.} $\mathrm{YTi_{0.332}Fe_{11.668}N_{0.5}}$ is a ThMn$_{12}$-type Fe-rich compound with dilute Fe/Ti substitution and partially occupied interstitial nitrogen sites. While the Ti occupancy is only about $8.3\%$ and the interstitial N site has $0.5$, both disorder channels are directly tied to permanent-magnet behavior through their effects on structural stability, Fe--Fe distances, and magnetocrystalline anisotropy. This case shows EP-Flow can capture the coupling between dilute substitutional disorder and interstitial N/vacancy disorder in a physically meaningful way.


\textbf{Mixed-anion perovskite with multi-sublattice substitution.} $\mathrm{Sr_{1.8}La_{0.2}FeCoO_{4.242}F_{1.758}}$ can be approximately written as ${\mathrm{(Sr_{0.9}La_{0.1})(Fe_{0.5}Co_{0.5})(O_{0.707}F_{0.293})_3}}$, corresponding to an oxyfluoride perovskite with simultaneous A-site Sr/La mixing, B-site Fe/Co mixing, and anion-site O/F mixing. This is a challenging multi-sublattice disorder case because both cation and anion sublattices are disordered, and the O/F arrangement directly affects local charge balance, covalency, and crystal-field splitting around the transition-metal centers. The example therefore tests whether EP-Flow can recover coupled mixed-cation and mixed-anion disorder within a single framework.


Taken together, these examples show that EP-Flow is not restricted to simple cation-site disorder: it recovers mixed-anion substitution in a perovskite, interstitial N/vacancy disorder in a Fe-rich magnetic framework, and oxygen-vacancy disorder in a layered superconducting oxide. Because these disorder patterns are directly linked to functional properties, successful reconstruction provides stronger evidence that the model captures realistic, property-relevant local chemistry. For reconstructions and analyses of more complex cases, please see \cref{app:structural_reconstruction}.

\subsubsection{Disorder-Pattern Analysis}

\begin{wrapfigure}{r}{0.48\textwidth}
\vspace{-1.2cm}
\centering
\includegraphics[width=0.50\textwidth]{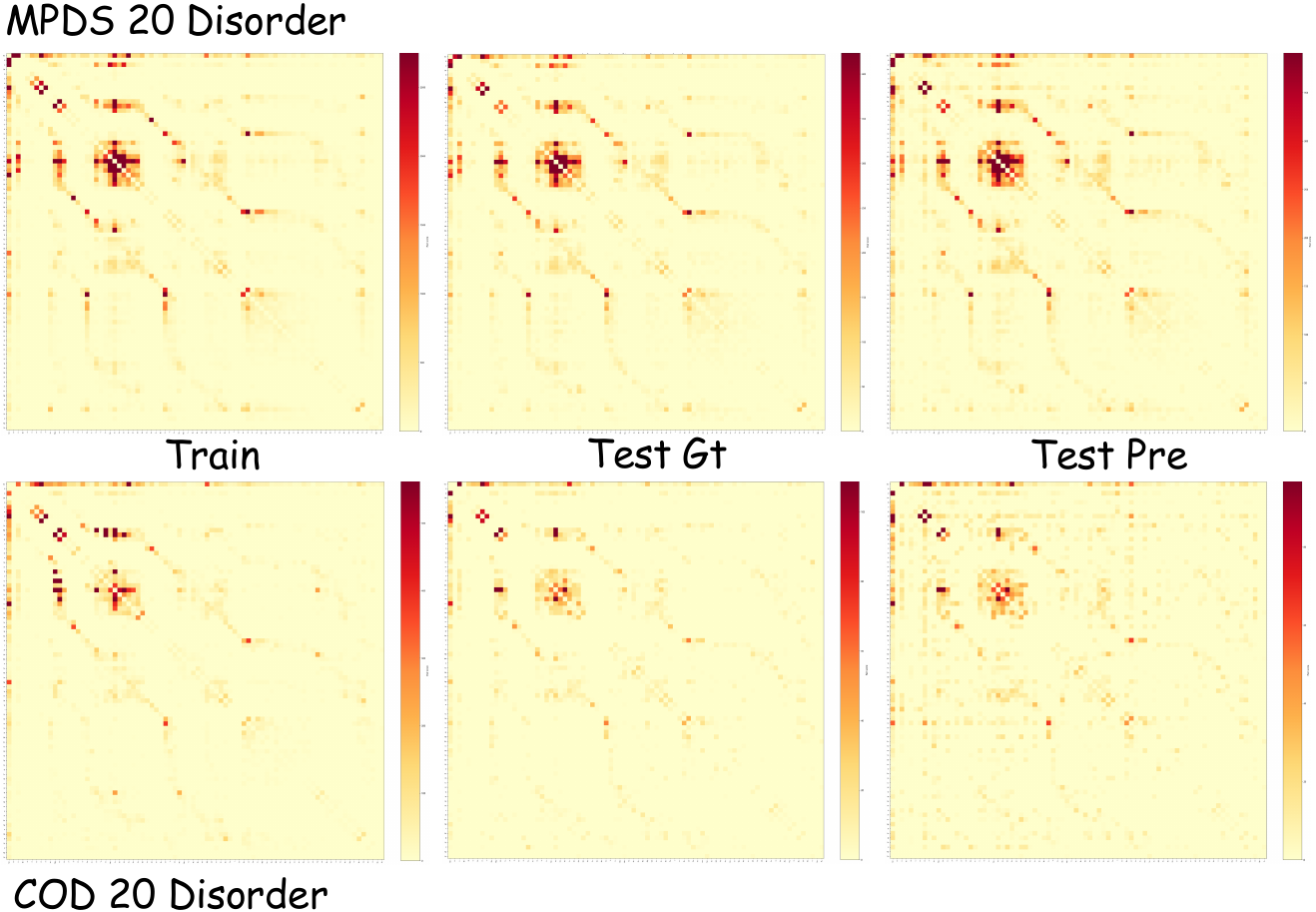}
\vspace{-0.45cm}
\caption{Site-wise occupancy-pattern analysis.}
\label{fig:analysis_heatmap}
\vspace{-0.15cm}
\end{wrapfigure}

To probe what the model learns, \cref{fig:analysis_heatmap} visualizes site-wise occupancy correlations. EP-Flow produces sparse, structured site--species assignments rather than diffuse occupancies: strong diagonal blocks indicate stable site preferences, while localized off-diagonal responses capture mixed-occupancy behavior. Moreover, some predicted mixing patterns absent from the smaller COD training split can still be verified in the larger MPDS dataset, suggesting that the model learns transferable site--species compatibility rather than merely memorizing observed occupancy pairs.

\section{Conclusion}\label{sec:conclusion}
We presented \ModelName, a geometric flow matching framework for disordered CSP conditioned on the formula and site number. A key contribution is formalizing this generation task to unify ordered structures with substitutional, vacancy, and interstitial disorder through the Occupancy Distribution Matrix (ODM). Methodologically, \ModelName canonicalizes heterogeneous, formula- and site-dependent occupancy polytopes into a shared, unbounded double-centered subspace. Here, it learns a marginal-preserving flow and maps samples back via Sinkhorn projection to effectively ensure physical validity. Together with a complete evaluation framework built from large-scale COD and MPDS data and metrics such as the Structure Match Rate tailored for disordered structures, this yields a unified pipeline for jointly generating occupancies, fractional coordinates, and lattice parameters directly from the given formula and site count. Experiments show that \ModelName achieves state-of-the-art performance across disorder types and site-count scales, highlighting the advantage of jointly modeling occupancy, geometry, and lattice under exact compositional constraints.

\bibliography{reference}
\bibliographystyle{unsrt}

\newpage

\appendix
\input{arxiv_appendix}


\newpage

\end{document}

%% file: arxiv_appendix.tex
\crefalias{section}{appendix}
\crefalias{subsection}{appendix}
\newcommand{\gMM}{\mathcal{M}_0(N,M)}

\section{Proofs and Foundational Concepts}\label{app:proof}

\subsection{Proof of the Bijectivity of Canonicalization}
\label{app:proof-bijection}

We restate the main claim using the notation of the paper. For a fixed stoichiometry vector $\vc \in \mathbb{R}_{>0}^{M}$ with total mass $N=\1^{\top}\vc$, define the transportation polytope
\begin{equation}
\gP(\1,\vc)
=
\left\{
\mP\in\mathbb{R}_{>0}^{N\times M}
\;\middle|\;
\mP\1_M=\1_N,\; \mP^{\top}\1_N=\vc
\right\},
\end{equation}
and the centered space

\begin{equation}
    \mathcal{M}_0(N,M)
    \triangleq 
    \Big\{
    \mZ\in\mathbb{R}^{N\times M}
    \;\big|\; 
    \mZ\bm{1}_M=\bm{0}_N,\; 
    \bm{1}_N^\top\mZ=\bm{0}_M^\top 
    \Big\}.
\end{equation}

\begin{proof}
We first prove surjectivity. Fix any $\mZ\in\gMM$ and define $\mB=\exp(\mZ)$ elementwise. Since $\mB$ is strictly positive, the Sinkhorn--Knopp theorem guarantees the existence of positive diagonal matrices $\mD_1$ and $\mD_2$ such that
\begin{equation}
\mA=\mD_1\mB\mD_2\in \gP(\1,\vc).
\end{equation}
We can explicitly express these strictly positive diagonal matrices using real-valued vectors $\va\in\mathbb{R}^{N}$ and $\vb\in\mathbb{R}^{M}$ as $\mD_1 = \mathrm{diag}(\exp(\va))$ and $\mD_2 = \mathrm{diag}(\exp(\vb))$. 

Substituting these into the scaling equation and taking the element-wise logarithm seamlessly translates the multiplicative Sinkhorn scaling into additive row and column shifts:
\begin{equation}
\log \mA = \mZ + \va\1^{\top} + \1\vb^{\top}.
\end{equation}
Applying the double-centering operator gives
\begin{equation}
\Psi(\mA)\triangleq\mathfrak{D}(\log \mA)=\mathfrak{D}(\mZ)=\mZ,
\end{equation}
because $\mathfrak{D}$ removes all additive row-wise and column-wise components and $\mZ$ is already centered. Hence every $\mZ\in\gMM$ has a preimage.

To prove injectivity, suppose $\mA_1,\mA_2\in\gP(\1,\vc)$ satisfy $\Psi(\mA_1)=\Psi(\mA_2)=\mZ$. Then both $\mA_1$ and $\mA_2$ are strictly positive matrices whose logarithms differ from $\mZ$ only by row-wise and column-wise additive terms. Exponentiating shows that both matrices are diagonal scalings of $\exp(\mZ)$ with the same target marginals $(\1,\vc)$. By uniqueness of Sinkhorn scaling for strictly positive matrices, these two scalings must coincide, so $\mA_1=\mA_2$.

Therefore $\Psi$ is bijective, and its inverse is exactly the Sinkhorn recovery
\begin{equation}
\Psi^{-1}(\mZ)=\gS(\exp(\mZ),\1,\vc).
\end{equation}
\end{proof}

\subsection{Trajectory Feasibility in the Centered Space}\label{app:proof-Trajectory}

We next justify why projecting the occupancy velocity onto $\gMM$ preserves the centered constraints along the entire ODE trajectory.

\begin{proof}
Let $\mA_t'\in\mathbb{R}^{N\times M}$ satisfy the ODE
\begin{equation}
\frac{d}{dt}\mA_t' = v_\theta^{\mA'}(\gD_t, t\mid\vc),
\qquad \mA_0'\in\gMM,
\end{equation}
and it satisfies $v_\theta^{\mA'}\in\gMM$ for all $t$. Consider the row-sum vector $\vr_t'=\mA_t'\1$ and the column-sum vector $\vc'_t=(\mA_t')^{\top}\1$. Differentiating yields
\begin{equation}
\frac{d}{dt}\vr_t' = \frac{d}{dt}\mA_t'\1 = v_\theta^{\mA'}(\gD_t, t\mid\vc)\1=\bm{0},
\end{equation}
and similarly
\begin{equation}
\frac{d}{dt}\vc'_t = \left(v_\theta^{\mA'}(\gD_t, t\mid\vc)\right)^{\top}\1=\bm{0}.
\end{equation}
Since $\mA_0'\in\gMM$, we have $\vr'_0=\bm{0}$ and $\vc'_0=\bm{0}$. Therefore $\vr_t'=\bm{0}$ and $\vc'_t=\bm{0}$ for every $t\in[0,1]$, which implies $\mA_t'\in\gMM$ for all $t$.
\end{proof}

Combining this fact with the inverse map $\Psi^{-1}$ immediately shows that the recovered trajectory $\mA_t=\gS(\exp(\mA_t'),\1,\vc)$ stays inside $\gP(\1,\vc)$ at all times.

\subsection{$SO(3)$ Invariance of the Joint Velocity Field}
\label{app:proof-so3}

For our model, a global rigid rotation acts only on the orientation of the lattice basis in Cartesian space. However, the network does not use the oriented basis $\mL_t$ directly. Instead, it takes the lattice parameters
$\vl_t=(a,b,c,\alpha,\beta,\gamma)$
as input, together with $\mA_t'$, $\mF_t$, and $\vc$. These six lattice parameters are invariant under any global rotation $Q\in SO(3)$, because such a rotation changes the embedding of the unit cell in Euclidean space but does not change its lengths or inter-axis angles. Therefore,
\begin{equation}
\vl_t(Q\mL_t)=\vl_t(\mL_t).
\end{equation}
At the same time, the fractional coordinates $\mF_t$ are intrinsic coordinates with respect to the lattice basis, so they are unchanged by a simultaneous rotation of the whole crystal. Hence the full model input is rotation-invariant. Since the network is a deterministic function of these rotation-invariant inputs, its outputs are also invariant. 

Therefore, a global $SO(3)$ rotation does not affect the predicted occupancy, coordinate, or lattice-parameter velocities. In this sense, the model is rotation-invariant at the level of its input representation and output predictions.

\subsection{Translation Invariance of the Joint Velocity Field}
\label{app:proof-translation}

We next consider periodic translations on the torus. It suffices to focus on the fractional coordinates $\mF_t$, since the other inputs $\mA_t'$, $\vl_t$, and $\vc$ are unaffected by a global shift of the crystal origin. Let $w(\cdot)$ denote the elementwise wrapping operator on the torus, and consider a periodic translation by $\bm{\tau}\in\mathbb{R}^{3}$:
\begin{equation}
\mF_t\mapsto w(\mF_t+\1\bm{\tau}^{\top}).
\end{equation}
Because the network only uses relative fractional displacements, for any edge $(i,j)$ we have
\begin{equation}
w\!\left(w(\vf_j+\bm{\tau})-w(\vf_i+\bm{\tau})\right)=w(\vf_j-\vf_i).
\end{equation}
Hence the periodic offset is unchanged under a common translation of all sites,
\begin{equation}
\Delta\vf_{ij}\big(w(\mF_t+\1\bm{\tau}^{\top})\big)=\Delta\vf_{ij}(\mF_t),
\end{equation}
and therefore its sinusoidal embedding is also unchanged. Since the remaining inputs are unchanged as well, the entire network input representation is translation-invariant.

\subsection{Flow Matching}
\label{app:flow_matching}
Flow Matching (FM)~\cite{albergobuilding,lipmanflow} defines a continuous-time transformation between a base distribution $p_0$ and a target data distribution $p_1$. Let $\vx_t \in \mathbb{R}^d$ denote a time-dependent data point along the probability path $p_t(\vx)$, where $t \in [0, 1]$. This path is governed by an Ordinary Differential Equation (ODE): $\frac{d}{dt} \vx_t = v_t(\vx_t)$, where $v_t$ is a time-dependent vector field. Since the true score or vector field of the marginal distribution $p_t$ is typically intractable, Conditional FM~\cite{lipmanflow} introduces a conditional probability path $p_t(\vx|\vx_1)$ that interpolates between a noise sample $\vx_0 \sim p_0$ and a data sample $\vx_1 \sim p_1$. Under a linear interpolation scheme, the intermediate state is defined as $\vx_t = t\vx_1 + (1-t)\vx_0$. The model is trained to approximate the ideal vector field by minimizing the expected $L_2$ distance between the learned vector field $v_\theta(\vx_t, t)$ and the tractable conditional vector field $v_t(\vx_t|\vx_1)$. Once trained, new samples are generated by numerically integrating the ODE from $t=0$ to $t=1$. 

Notably, when data are naturally constrained to specific geometric structures, such as the torus or the simplex, standard Euclidean FM is extended to Riemannian FM~\cite{chen2024rfm}. This framework defines the probability path and vector field directly on the manifold's tangent bundle, ensuring that the generative trajectory respects the underlying geometric boundaries.

\clearpage

\section{Physical Intuition and Conceptual Inspirations}
\label{app:physical_interpretations}

The algebraic operations of the canonicalization map $\Psi$ and its inverse can be understood as a \textbf{conceptual analogy} to the principles of statistical thermodynamics governing continuous atomic distributions. We detail the physical mappings of the four main modeling components below.

\subsection{Logarithmic Mapping as Energy Space Evolution}
In our continuous formulation, the fractional occupancy $A_{ij} \in (0,1)$ represents the probability of species $j$ occupying site $i$. This suggests a \textbf{Boltzmann-type interpretation}, where the occupancy is conceptually related to a latent local formation energy $E_{ij}$ via:
\begin{equation}
A_{ij} = \frac{1}{Q} \exp\left(-\frac{E_{ij}}{k_B T}\right),
\end{equation}
where $Q$ is the partition function and $T$ serves as an effective temperature. Taking the element-wise logarithm yields:
\begin{equation}
\log A_{ij} = -\frac{E_{ij}}{k_B T} - \log Q.
\end{equation}
By mapping the bounded probability simplex into an unconstrained energy space, the generative ODE avoids learning a velocity field over restricted probability boundaries. Consequently, the vector field learned by the model can be viewed as an analogy to the rate of change of an energy landscape with respect to generative time ($\propto -\partial E_{ij} / \partial t$). This enables the network to simulate a continuous thermodynamic-like evolution, driving the system from high-entropy noise toward a low-energy configuration. Note that this formulation characterizes the generative trajectory as an evolution toward potential stability, rather than assuming the system maintains global thermodynamic equilibrium at every intermediate step.

\subsection{Double-Centering as Relative Potential Extraction}
In thermodynamics, absolute energy scales are unobservable; system dynamics are driven exclusively by relative energy differences. We decompose the effective local formation energy $E_{ij}$ into a centered core interaction and macroscopic background fields:
\begin{equation}
E_{ij} = E_{ij}^{\text{core}} + \mu_i^{\text{site}} + \mu_j^{\text{element}} + E_0.
\end{equation}
Here, $\mu_i^{\text{site}}$ is the site-specific potential well, $\mu_j^{\text{element}}$ is the species-specific chemical potential determined by stoichiometry $\vc$, and $E_{ij}^{\text{core}}$ is the intrinsic relative energy. Applying the double-centering operator $\mathfrak{D}$ to $\log \mA$ \textbf{algebraically eliminates} these additive background fields and constant terms, leaving:
\begin{equation}
Z_{ij} = \mathfrak{D}(\log \mA)_{ij} = -\frac{E_{ij}^{\text{core}}}{k_B T}.
\end{equation}
Therefore, $\mZ$ represents a dimensionless relative affinity, acting as an \textbf{intrinsic structural invariant} of the local chemical bonding. Operating in $\mathcal{M}_0(N,M)$ therefore preserves the information needed to recover the ODM under fixed marginals, while providing an energy-like representation for learning.

\subsection{Sinkhorn Recovery as Thermodynamic Annealing}
The inverse mapping reconstructs the physical occupancy $\mA$ by applying the Sinkhorn algorithm to $\exp(\mZ)$, converging to:
\begin{equation}
A_{ij} = \exp(\va_i) \exp\left(-\frac{E_{ij}^{\text{core}}}{k_B T}\right) \exp(\vb_j).
\end{equation}
In the context of a grand canonical ensemble, the scaling variables $\va_i$ and $\vb_j$ function as dynamic chemical potentials. The iterative Sinkhorn scaling provides a mathematical \textbf{analogy} to an isothermal annealing process where atomic distributions are adjusted to satisfy global stoichiometric constraints. Structural formation is finalized when the system reaches a state that \textbf{effectively satisfies} the physical boundary constraints imposed by the target formula.

\subsection{Physical Interpretation of Energy-Difference Edge Features}\label{app:Energy-Differenc}
In the message-passing network, we include the difference between connected sites in the double-centered log-domain representation:
\begin{equation}
    \Delta \mA'_{ij} = \mA'_{i,:} - \mA'_{j,:}.
\end{equation}
This feature is computed from the centered representation $\mA'=\mathfrak{D}(\log \mA)$, which removes background terms and can be interpreted as a relative affinity field. This log-domain difference has a direct probabilistic interpretation: for a given species $e$, a difference in log occupancy corresponds to a probability ratio $\log (A_{i,e}/A_{j,e})$. Under the Boltzmann-type interpretation, this ratio further corresponds to a relative energy difference $-(E_{i,e}-E_{j,e})/k_BT$. Subtraction in the log-domain space thus encodes relative site preferences. By providing the network with explicit information about how the relative energy landscape changes across the crystal graph, $\Delta \mA'_{ij}$ helps the model distinguish locally compatible site--species configurations. The ablation results in \cref{tab:ablation} confirm that this feature is critical for recovering accurate local disorder patterns.

\vspace{1em}
\noindent \textbf{Concluding Remarks on Physical Analogy.} This conceptual bridge between manifold-constrained flow and statistical mechanics represents an \textbf{exploratory attempt} to provide physical intuition for disordered crystal generation. While these analogies are intended as a heuristic framework rather than a formal derivation, they offer a new vantage point for designing constraint-aware generative models in materials science. We hope this perspective stimulates further cross-disciplinary exploration between the machine learning and physical science communities.

\section{Data Preprocessing and Metrics}\label{app:data_metric}

\subsection{Datasets}\label{sec:datasets_supp}
We evaluate the model on four subsets derived from the COD and MPDS databases: substitutional-disorder and general-disorder settings, each at two scales (up to 20 and 50 sites). Following the protocol in the main text, each subset is split into training, validation, and test sets with an approximate ratio of $8{:}1{:}1$. All models are trained conditionally on the stoichiometry vector $\vc$.
Crucially, to provide a unified thermodynamic representation, we treat crystallographic vacancies as an explicit pseudo-atom. Consequently, the stoichiometry vector $\vc$ includes this vacancy species, ensuring that the total sum of the formula strictly equals the number of crystallographic sites ($\1^{\top}\vc = N$).

All raw CIF files are standardized by primitive-cell reduction followed by Niggli reduction.

We then apply a set of filters to ensure physical validity, computational tractability, and sufficient statistical support. 
Structures with interatomic distances below 0.5\,\AA, cell volume below 0.1\,\AA$^3$, or any lattice parameter larger than 40\,\AA\ are removed. 

To evaluate different structural scales, we construct subsets with the number of crystallographic sites satisfying $N\leq 20$ or $N\leq 50$.

To control the compositional complexity, we limit the number of distinct species in each extended formula to $M\leq 6$, including the vacancy pseudo-atom when present.

To focus on inorganic crystals and avoid molecular orientational disorder, we exclude structures containing both C and H.

To restrict local disorder complexity, we allow at most four co-occupying species on each mixed site and require each retained structure to contain at least one disordered site.

To ensure statistically reliable elemental embeddings, we remove structures containing rare elements, defined as elements appearing fewer than 5 times in raw COD or fewer than 50 times in raw MPDS.

\begin{figure}[t]
\vspace{-0.2cm}
    \centering
    \includegraphics[width=\linewidth]{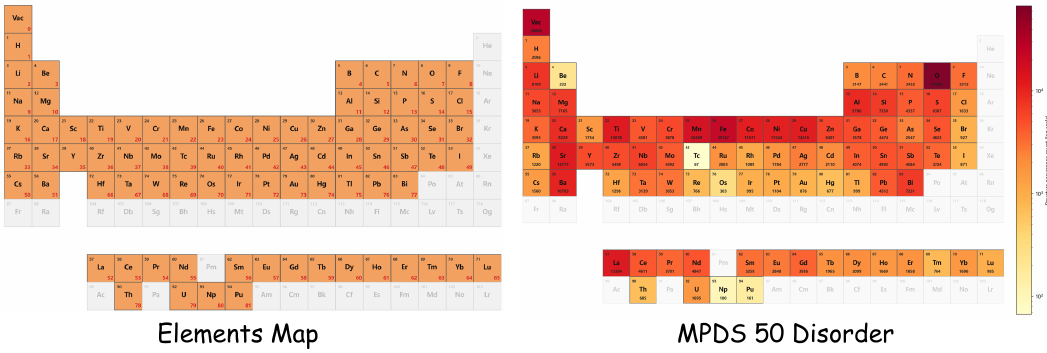}
    \vspace{-0.5cm}

    \caption{
    \textbf{Element index map and MPDS Disorder-50 distribution.}
    Left: the element-to-position index map used in this work.
    Right: the element-occurrence heatmap for the MPDS Disorder-50 subset.
    }
    \label{fig:element_map}
    \vspace{-0.2cm}

\end{figure}

Figure~\ref{fig:element_map} illustrates the element-to-position index map used in this work and the resulting element-occurrence distribution in the MPDS Disorder-50 subset.

To prevent data leakage during evaluation, we apply a strict structural deduplication protocol before the train/validation/test split. Structures with matching formulas and site counts are compared using structure matchers with
\begin{equation}
(\mathrm{ltol},\mathrm{stol},\mathrm{angle\_tol},\mathrm{occtol})
=
(0.2,0.3,5,0.1).
\end{equation}
Duplicates are rigorously removed from the validation and test sets.

Table~\ref{tab:dataset-stats} reports the size of each de-duplicated subset. The Sub-20/50 subsets contain substitutional disorder only, whereas Disorder-20/50 additionally include vacancy disorder and therefore represent a more challenging generation regime.
\begin{table}[t]
\centering
\caption{Dataset statistics for COD and MPDS databases.}
\label{tab:dataset-stats}
\begin{tabular}{llrrrrr}
\toprule
\textbf{Database} & \textbf{Setting} & \textbf{Train} & \textbf{Val} & \textbf{Test} & \textbf{Total} & \textbf{Ratio (Train:Val:Test)} \\
\midrule

\multirow{4}{*}{\textbf{COD}} 
& Substitution 20 & 7,100  & 879   & 921   & 8,900   & $7.98 : 0.99 : 1.03$ \\
& Substitution 50 & 9,483  & 1,141 & 1,190 & 11,814  & $8.03 : 0.97 : 1.01$ \\
& Disorder 20     & 9,408  & 1,185 & 1,217 & 11,810  & $7.97 : 1.00 : 1.03$ \\
& Disorder 50     & 13,825 & 1,689 & 1,725 & 17,239  & $8.02 : 0.98 : 1.00$ \\

\midrule

\multirow{4}{*}{\textbf{MPDS}} 
& Substitution 20 & 53,492 & 6,648  & 6,786  & 66,926  & 7.99 : 0.99 : 1.01 \\
& Substitution 50 & 65,187 & 8,412  & 8,499  & 82,098  & $7.94 : 1.02 : 1.03$ \\
& Disorder 20     & 69,032 & 8,715  & 8,835  & 86,582  & $7.97 : 1.01 : 1.02$ \\
& Disorder 50     & 87,882 & 11,531 & 11,553 & 110,966 & $7.92 : 1.04 : 1.04$ \\

\bottomrule
\end{tabular}
\end{table}



\subsection{Baseline Adaptations}
\label{app:baseline_adaptations}

\paragraph{Ordered Baselines.}
Since FlowMM~\cite{miller2024flowmm} and OMatG~\cite{hollmer2025omatg} are originally designed for ordered crystal structure prediction, we adapt them to support continuous site-wise occupancies while preserving their original backbone architectures. Specifically, we replace the discrete atomic-species representation with continuous occupancy distributions and introduce an additional prediction head to jointly generate site occupancies with fractional coordinates and lattice parameters. The input chemical formula is used to mask species absent from the target composition, and a Sinkhorn projection is applied to the predicted occupancy matrix at inference time to enforce the target composition and site-wise
normalization constraints. For OMatG, we additionally incorporate the
disordered formula embedding used in EP-Flow, which provides explicit
composition and site information to the network. Other components of the
original FlowMM and OMatG architectures are kept unchanged.

\paragraph{Disordered Baselines.}
The original CSP setting of DMFlow~\cite{wu2026dmflow} assumes that site-wise occupancy distributions are provided as input and remain fixed during generation, which is incompatible with our formula-conditioned setting where the occupancies are unknown. We therefore adapt the occupancy-generation mechanism from DMFlow's De Novo Generation (DNG) setting to predict site-wise occupancies from the chemical formula. In particular, we employ its Riemannian spherical reparameterization to constrain the generated occupancy distributions, while jointly generating the structural variables using the original DMFlow framework. At inference time, we further apply a Sinkhorn projection to the predicted occupancies to enforce consistency with the target chemical formula. Thus, Adapted-DMFlow does not use ground-truth site-level occupancy information as input and receives the same task-level conditioning information as EP-Flow.

\paragraph{Summary of constraint handling.}
These adaptations also provide a progressive comparison of how occupancy
constraints are incorporated into the generative process. Adapted-FlowMM and Adapted-OMatG generate occupancies without explicitly enforcing either the row or column constraints during generation, and rely on a post-hoc Sinkhorn projection to satisfy both constraints. Adapted-DMFlow introduces the site-wise row constraint through its spherical reparameterization, such that each site represents a normalized occupancy distribution, but does not enforce the column constraint corresponding to the target composition during generation; the latter is again imposed through the Sinkhorn post-processing. In contrast, EP-Flow explicitly preserves both the row and column constraints throughout the generative process. Therefore, these baselines form a progressive comparison from unconstrained generation, to row-constrained
generation, and finally to the jointly row- and column-constrained generation of EP-Flow, allowing us to directly assess the benefit of incorporating the full transportation-polytope constraints into the generative dynamics.

\subsection{Evaluation Metrics}\label{sec:metric_supp}
To rigorously evaluate disordered crystals, we propose an extended structure matching metric suite. Building upon the standard evaluation protocol for ordered crystals, we additionally incorporate continuous site-wise occupancies into the structural matching process. Based on this, our models are evaluated using four key metrics: Match Rate, RMSD, OccMAE, and structural validity.

Let $\{(\hat{\mathcal{D}}^{(n)},\mathcal{D}^{(n)})\}_{n=1}^{N_{\mathrm{test}}}$ denote the generated/reference pairs. The structure-matching stage uses
\begin{equation}
(\mathrm{ltol},\mathrm{stol},\mathrm{angle\_tol},\mathrm{occtol})=(0.3,0.5,10,0.3),
\end{equation}
where $\mathrm{ltol}$ is the relative lattice-length tolerance, $\mathrm{stol}$ is the normalized site-distance tolerance, $\mathrm{angle\_tol}$ is the lattice-angle tolerance in degrees, and $\mathrm{occtol}$ is the occupancy tolerance.

The choice of $\mathrm{occtol}=0.3$ is designed to align seamlessly with the relative scale of the lattice and spatial tolerances ($\mathrm{ltol}=0.3, \mathrm{stol}=0.5$). This threshold permits reasonable generative variance in continuous fractional occupancies while rigorously rejecting fundamentally mismatched atomic assignments. 

To demonstrate that this threshold does not artificially inflate performance, we comprehensively evaluated EP-Flow on the representative MPDS Sub-20 dataset under increasingly stringent occupancy tolerances, ranging from 0.30 down to a highly strict 0.01. 

As summarized in \cref{tab:occtol_robustness}, tightening $\mathrm{occtol}$ from 0.30 to 0.01 only reduces the Match Rate from 60.82\% to 50.09\%, while the OccMAE naturally decreases to 0.0004. The robust retention of the Match Rate under such extreme constraints confirms that the model genuinely captures highly precise continuous site-wise occupancies, rather than merely passing a loose matching threshold.

\begin{table}[htbp]
\centering
\caption{Performance of EP-Flow on the MPDS Sub-20 dataset under varying occupancy tolerances.}
\label{tab:occtol_robustness}
\begin{tabular}{ccccc}
\toprule
$\mathrm{occtol}$ & Successful Matches & Match Rate & Normalized RMSD & OccMAE \\
\midrule
0.30 & 4,127 & 60.82\% & 0.0483 & 0.0075 \\
0.25 & 4,067 & 59.93\% & 0.0483 & 0.0058 \\
0.20 & 4,012 & 59.12\% & 0.0480 & 0.0048 \\
0.15 & 3,916 & 57.71\% & 0.0480 & 0.0033 \\
0.10 & 3,804 & 56.06\% & 0.0479 & 0.0020 \\
0.05 & 3,653 & 53.83\% & 0.0475 & 0.0010 \\
0.01 & 3,399 & 50.09\% & 0.0444 & 0.0004 \\
0.005 & 3,186 & 46.95\% & 0.0438 & 0.0002 \\
\bottomrule
\end{tabular}
\end{table}

For each pair, we first compute a vacancy-aware site-wise occupancy discrepancy matrix. Let $p_{i,e}^{(n)}$ and $g_{j,e}^{(n)}$ denote the occupancies of species $e$ at predicted site $i$ and reference site $j$, respectively, with vacancy included explicitly as an additional species. Then
\begin{equation}
d_{\mathrm{occ}}^{(n)}(i,j)
=
\frac{1}{2}
\sum_{e}
\left|p_{i,e}^{(n)}-g_{j,e}^{(n)}\right|,
\end{equation}
which is exactly the total variation distance. A site pair is chemically feasible if $
d_{\mathrm{occ}}^{(n)}(i,j)\le \mathrm{occtol}$.

Next, for every lattice-aligned candidate and periodic shift, the pairwise spatial distance is computed on the ground-truth lattice and normalized by the average free length
\begin{equation}
\ell_{\mathrm{free}}^{(n)}=\left(\frac{\mathrm{Vol}(\mathcal{D}^{(n)})}{N_n}\right)^{1/3},
\qquad
d_{\mathrm{spa}}^{(n)}(i,j)=\frac{\mathrm{dist}^{(n)}(i,j)}{\ell_{\mathrm{free}}^{(n)}},
\end{equation}
where $\mathrm{Vol}(\cdot)$ denotes the volume and $\mathrm{dist}(\cdot,\cdot)$ is the distance between two atoms. A site pair is spatially feasible if $d_{\mathrm{spa}}^{(n)}(i,j)\le \mathrm{stol}$.

The final valid edge set is the intersection of the occupancy-feasible and spatially feasible masks. In this bipartite graph, we solve a Hungarian assignment~\cite{kuhn1955hungarian} by minimizing the normalized spatial cost. A sample is counted as a successful match only when a perfect $N_n$-to-$N_n$ assignment exists.

Let $\mathcal{V}=\{n\mid \hat{\mathcal{D}}^{(n)}\ \text{and}\ \mathcal{D}^{(n)}\ \text{are both valid 3D structures}\}$, $\pi_n$ denote the optimal assignment for sample $n$, and
\begin{equation}
\mathcal{S}=\left\{n\in\{1,\dots,N_{\mathrm{test}}\}\;\middle|\; \text{sample $n$ admits a valid perfect assignment}\right\},
\end{equation}
be the set of successful matches. In the implementation, $\mathcal{S}\subseteq \mathcal{V}$ because the matching stage is evaluated only when both the prediction and the ground truth pass the validity checks. Then the \textbf{Match Rate} reported by the code is
\begin{equation}
\mathrm{MR}=\frac{|\mathcal{S}|}{N_{\mathrm{test}}}.
\end{equation}
For each successful sample $n\in\mathcal{S}$, the per-structure normalized RMSD is
\begin{equation}
\mathrm{RMSD}^{(n)}
=
\sqrt{\frac{1}{N_n}\sum_{i=1}^{N_n} \left(d_{\mathrm{spa}}^{(n)}\big(i,\pi_n(i)\big)\right)^2},
\end{equation}
and the reported dataset-level \textbf{RMSD} is the average over successful samples,
\begin{equation}
\mathrm{RMSD}=\frac{1}{|\mathcal{S}|}\sum_{n\in\mathcal{S}} \mathrm{RMSD}^{(n)}.
\end{equation}
Similarly, the per-structure occupancy error is
\begin{equation}
\mathrm{OccMAE}^{(n)}
=
\frac{1}{N_n}\sum_{i=1}^{N_n} d_{\mathrm{occ}}^{(n)}\big(i,\pi_n(i)\big),
\end{equation}
and the reported \textbf{OccMAE} is
\begin{equation}
\mathrm{OccMAE}=\frac{1}{|\mathcal{S}|}\sum_{n\in\mathcal{S}} \mathrm{OccMAE}^{(n)}.
\end{equation}

The \textbf{Validity} score follows the crystal-construction and sanity checks: a prediction is valid only if it can be converted to a pymatgen structure~\cite{ong2013pymatgen} and passes the structural validity test. Therefore,
\begin{equation}
\mathrm{Validity}
=
\frac{1}{N_{\mathrm{test}}}
\sum_{n=1}^{N_{\mathrm{test}}}
\mathbb{I}\!\left[\hat{\mathcal{D}}^{(n)}\ \text{has no structure overlap}\right].
\end{equation}


\clearpage

\section{Implementation Details}\label{app:imple_details}

\subsection{Disordered Crystal Structure}\label{app:details_disorder}
\begin{figure}[htbp]
  \centering
  \includegraphics[width=0.9\textwidth]{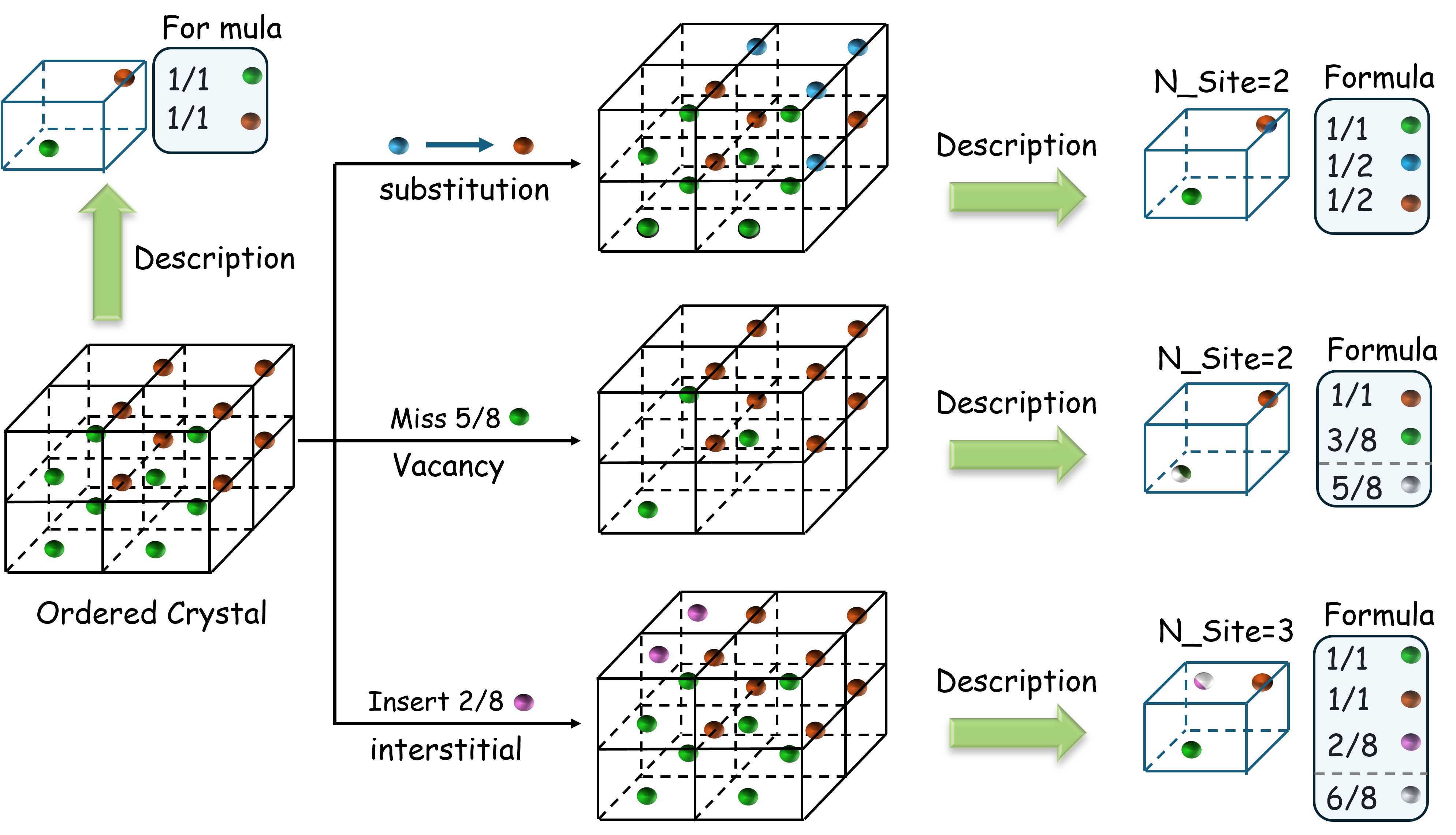} 
  \caption{
    \textbf{Unified representation of diverse disorder types via the vacancy-extended Occupancy Distribution Matrix (ODM).} 
    In an ordered crystal (left), the site count $N$ is well-defined by the formula with discrete occupancies. When transitioning to disordered systems (right), this assumption breaks down. For \textbf{substitutional disorder} (top right), occupancies become continuously fractional while $N$ remains unchanged. For \textbf{vacancy} (middle right) and \textbf{interstitial disorder} (bottom right), the site count cannot be determined from the raw formula alone. To resolve this ambiguity, a vacancy pseudo-atom ($\mathrm{Vac}$, depicted as grey/white spheres) is introduced to ensure that the occupancy at each site sums to one. This yields an extended formula $\mathbf{c}$ that incorporates the total vacancy count, satisfying $\sum_{j=1}^{M} c_j = N$. This formulation explicitly conditions the structural representation on both the formula and the site number, allowing all disorder types to be modeled within a unified ODM framework.
  }
  \label{fig:odm_unification}
\end{figure}

Crucially, the Occupancy Distribution Matrix (ODM) should not be interpreted as a single, static snapshot of a disordered crystal. Instead, it serves as a statistical summary of all possible local atomic arrangements. As illustrated in Fig.~\ref{fig:odm_unification}, while an ordered structure assigns exactly one specific element to each crystallographic site, a disordered structure allows equivalent sites to host different elements---or remain completely empty---across different regions of the material. The vacancy-extended ODM captures this variability by assigning each site a probability distribution over all candidate species, explicitly including the vacancy pseudo-atom. Therefore, each entry $A_{ij}$ simply represents the probability, or the average fraction, that species $j$ occupies site $i$. This formulation enables \ModelName to model disordered crystals at an average, macroscopic level. It elegantly translates the physical complexity of mixed atomic arrangements into clear geometric marginal constraints, while still preserving physically meaningful, site-specific disorder patterns.
\subsection{Formula and Number of Sites}\label{app:details_for_site}

In ordered crystal generation, the number of atomic sites is usually well defined from the reduced chemical formula. 
Each crystallographic site is occupied by a single atom, and therefore the total number of atoms in the formula naturally equals the number of nodes in the crystal graph. 
This convention is adopted by many graph-based crystal generation models, such as DiffCSP and FlowMM, where the model typically needs to know or sample the number of nodes before generating fractional coordinates and lattice parameters.

However, this assumption does not directly hold for disordered crystals. 
In a disordered structure, a crystallographic site may be fractionally occupied by multiple species, or may be partially vacant. 
As a result, the raw chemical formula may contain non-integer stoichiometric coefficients and its total sum does not necessarily reveal the number of crystallographic sites. 
For example, a partially occupied site with occupancy $0.5$ contributes only half an atom to the chemical formula, while it still corresponds to one crystallographic site in the structure representation. 
Therefore, the site count cannot be inferred by simply summing the non-vacancy species in the formula.

To obtain a consistent representation, we explicitly introduce a vacancy species, denoted as $\mathrm{Vac}$, to complete each crystallographic site to full occupancy. 
For each site $i$, the occupancy vector satisfies
\begin{equation}
    \sum_{j=1}^{M} A_{ij} = 1,
\end{equation}
where $M$ includes both real chemical species and the vacancy species. 
Accordingly, the extended stoichiometric vector $\vc$ also includes the total vacancy count. 
This gives
\begin{equation}
    \sum_{j=1}^{M} c_j = N,
\end{equation}
where $N$ is the number of crystallographic sites. 
In other words, after vacancy completion, the extended formula is fully occupied at the site level, and the sum of the extended formula equals the number of sites.

To better illustrate how the vacancy-augmented formula bridges macroscopic chemical formulas with microscopic site occupancies, let us consider a common macroscopic formula, $A_{1.5}B_{1}$.

\noindent \textbf{Case 1 (A structure with 3 sites):} Suppose we read the structural information of a crystal with this formula and find it has exactly $N = 3$ crystallographic sites. The actual site occupancies might be: Site 1 (A, occupancy = 1.0), Site 2 (B, occupancy = 1.0), and Site 3 (A, occupancy = 0.5). To address the partial occupancy on Site 3, we introduce a ``vacancy'' ($\mathrm{Vac}$) pseudo-atom to pad the remaining probability. Thus, Site 3 becomes a mixed site: (A: 0.5, $\mathrm{Vac}$: 0.5). This transforms the vacancy defect into a purely substitution-disorder format. The corresponding augmented formula becomes $A_{1.5}B_{1}\mathrm{Vac}_{0.5}$. The sum of the base formula subscripts ($1.5 + 1 + 0.5 = 3$) perfectly matches the actual number of sites ($N=3$).

\noindent \textbf{Case 2 (A structure with 4 sites):} Now, suppose another structure with the exact same conventional formula $A_{1.5}B_{1}$ has $N = 4$ crystallographic sites. Its actual site occupancies might be: Site 1 (A, occupancy = 1.0), Site 2 (B, occupancy = 1.0), Site 3 (A, occupancy = 0.25), and Site 4 (A, occupancy = 0.25). We pad the empty probabilities on Sites 3 and 4 with the vacancy pseudo-atom, resulting in Site 3 (A: 0.25, $\mathrm{Vac}$: 0.75) and Site 4 (A: 0.25, $\mathrm{Vac}$: 0.75). The augmented formula for this specific structure becomes $A_{1.5}B_{1}\mathrm{Vac}_{1.5}$. Again, the sum of the base formula subscripts ($1.5 + 1 + 1.5 = 4$) perfectly matches the site count ($N=4$).

In practice, this simple yet effective padding mechanism guarantees that our transportation-polytope framework can unify substitution disorder and vacancy defects under a single mathematical formulation. During training, we utilize these fully padded, vacancy-augmented formulas alongside their corresponding ODMs. During inference, the augmented formula serves as the deterministic conditional input. It is directly derived from the conventional chemical formula and the target number of sites $N$ by calculating the vacancy amount as:
\begin{equation}
    \text{Amount}_{\mathrm{Vac}} = N - \sum(\text{base formula subscripts}).
\end{equation}
For our examples, this is $3 - (1.5 + 1) = 0.5$ for Case 1, and $4 - (1.5 + 1) = 1.5$ for Case 2.

This treatment is important for disordered crystal generation. 
It allows substitutional disorder and vacancy disorder to be represented in a unified occupancy distribution matrix (ODM), while preserving the same node-level structure as graph-based crystal models. 
The model therefore generates a fixed number of crystallographic sites $N$, each associated with a fractional coordinate and a probability distribution over real species and vacancies. 
The final physical composition is recovered by marginalizing out the vacancy species, whereas the extended formula ensures that every site remains fully occupied in the ODM representation.

\newpage

\subsection{Log-domain Double Centering}\label{alg:log_dc}

\begin{wrapfigure}{r}{0.40\textwidth}
\vspace{-0.5cm}
\centering
\begin{minipage}{0.4\textwidth}
\textbf{Log-domain Double Centering}
\footnotesize
\begin{algorithmic}[1]

\Ensure Centered representation $\mA'$
\State $\mU_{\mB}\gets \mathrm{MaskedUniform}(\mB)$
\State $\mA^{\epsilon}\gets (1-\epsilon)\mA+\epsilon\mU_{\mB}$
\State $\mX\gets \log \mA^{\epsilon}$
\State $\widetilde{\mX}\gets \mX\odot\mB$
\State $r\gets \mathrm{MaskedRowMean}(\widetilde{\mX},\mB)$
\State $q\gets \mathrm{MaskedColumnMean}(\widetilde{\mX},\mB)$
\State $g\gets \mathrm{MaskedGlobalMean}(\widetilde{\mX},\mB)$
\State $\mA'\gets \mB\odot(\widetilde{\mX}-r-q+g)$
\State \Return $\mA'$
\end{algorithmic}
\end{minipage}
\vspace{-0.2cm}
\end{wrapfigure}

\paragraph{\texorpdfstring{$\epsilon$}{epsilon}-smoothing and log-domain canonicalization.}
Real ODMs are highly sparse, and many entries of $\mA$ are exactly zero.
Directly applying $\log \mA$ is therefore ill-defined.
We first apply a small $\epsilon$-smoothing to move the ODM from the boundary of the transportation polytope to its relative interior:
\begin{equation}
\mA^{\epsilon} = (1-\epsilon)\mA+\epsilon\mU_{\mB},
    \qquad
    U_{ij} = \frac{B_{ij}}{\sum_{k=1}^{M}B_{ik}} .
\end{equation}

We then perform masked log-domain double centering:
\begin{equation}
    \mA'
    =
    \mathfrak{D}_{\mB}
    \left(
    \log \mA^{\epsilon}
    \right).
\end{equation}
This removes site-wise and species-wise additive effects and provides a shared zero-marginal representation for flow matching.

\subsection{Sinkhorn Recovery}\label{alg:sinkhorn_recovery}

\begin{wrapfigure}{r}{0.40\textwidth}
\vspace{-0.4cm}
\centering
\begin{minipage}{0.40\textwidth}
{\textbf{Sinkhorn Recovery}}
\footnotesize
\begin{algorithmic}[1]
\Ensure Recovered ODM $\mA$
\State $\mB\gets \mathbf{1}[\vc>0]$
\State $\mZ\gets \mathrm{MaskInvalid}(\mA',\mB)$
\State $\mZ\gets \mZ-\mathrm{RowMax}(\mZ)$
\State $\mK\gets \exp(\mZ)\odot\mB+\epsilon\mB$
\State $\mK\gets \mathrm{Sinkhorn}(\mK;\bm{1}_N,\vc,K)$
\State $K_{ij}\gets 0,\quad \text{if }0<K_{ij}<\tau$
\State $\mA\gets \mathrm{Sinkhorn}(\mK;\bm{1}_N,\vc,K)$
\State \Return $\mA$
\end{algorithmic}
\end{minipage}
\vspace{-0.2cm}
\end{wrapfigure}

\paragraph{Formula-conditioned recovery with truncation.}

The generated centered matrix $\mA'$ is not itself a valid occupancy matrix.
We first mask species absent from the target formula and exponentiate the stabilized logits to obtain a positive kernel:
\begin{equation}
    \mK
    =
    \exp(\mZ)\odot\mB+\epsilon\mB,
    \qquad
    \mB=\mathbf{1}[\vc>0].
\end{equation}
The kernel is then projected onto the formula-conditioned transportation polytope by Sinkhorn normalization:
\begin{equation}
    \mA
    =
    \mathcal{S}
    \left(
    \mK,
    \bm{1}_N,
    \vc
    \right),
\end{equation}
Specifically, the Sinkhorn algorithm seeks two positive scaling vectors $\vu \in \mathbb{R}_{>0}^N$ and $\vv \in \mathbb{R}_{>0}^M$ such that the scaled matrix $\mA = \mathrm{diag}(\vu) \mK \mathrm{diag}(\vv)$ meets the prescribed marginals. This is efficiently achieved through alternating multiplicative updates:
\begin{equation}
    \vu^{(k+1)} = \frac{\bm{1}_N}{\mK \vv^{(k)}}, \qquad \vv^{(k+1)} = \frac{\vc}{\mK^{\top} \vu^{(k+1)}},
\end{equation}
where the division is performed element-wise, and $\vv^{(0)}$ is typically initialized to $\bm{1}_M$. After a finite number of iterations $K_
\mathrm{SK}$, the recovered matrix $\mA = \mathrm{diag}(\vu^{(K)}) \mK \mathrm{diag}(\vv^{(K)})$ satisfies the marginal constraints up to numerical precision:\begin{equation}
    \mA\bm{1}_{M}=\bm{1}_N,
    \qquad
    \mA^{\top}\bm{1}_N=\vc,
    \qquad
    A_{ij}\geq 0 .
\end{equation}

During CIF reconstruction, we observe that the recovered ODM may contain very small nonzero occupancies on some site--species pairs with negligible contribution, sometimes around $10^{-5}$.
If directly written into CIF files, these trace values may be interpreted as additional mixed species, leading to unnecessarily dense occupancy annotations.

This behavior is mainly caused by the numerical recovery procedure.
First, Sinkhorn recovery operates on strictly positive kernels induced by $\epsilon$-smoothing, so valid entries may retain small positive masses after normalization.
Second, Sinkhorn normalization is implemented with a finite number of iterations, which can leave tiny residual values on numerically feasible site--species pairs.

To obtain cleaner reconstructed structures, we truncate entries with $0<A_{ij}<\tau$ to zero after the first Sinkhorn projection.
In our experiments, we set $\tau=0.005$, corresponding to occupancies below $0.5\%$ of a crystallographic site.
This threshold is intentionally conservative: it removes only trace values below the occupancy scale considered meaningful in our reconstruction protocol, while preserving genuine fractional occupancies.
After truncation, we apply a second Sinkhorn projection to restore the prescribed row and column marginals up to numerical tolerance.

This post-processing produces a sparse and formula-consistent ODM, helps avoid interpreting negligible numerical residuals as meaningful mixed-site disorder during CIF reconstruction, and yields reconstructed crystals with more physically interpretable occupancies.

\subsection{Structural Reconstruction}
\label{app:structural_reconstruction}

\begin{figure}[h]
    \centering
    \includegraphics[width=\linewidth]{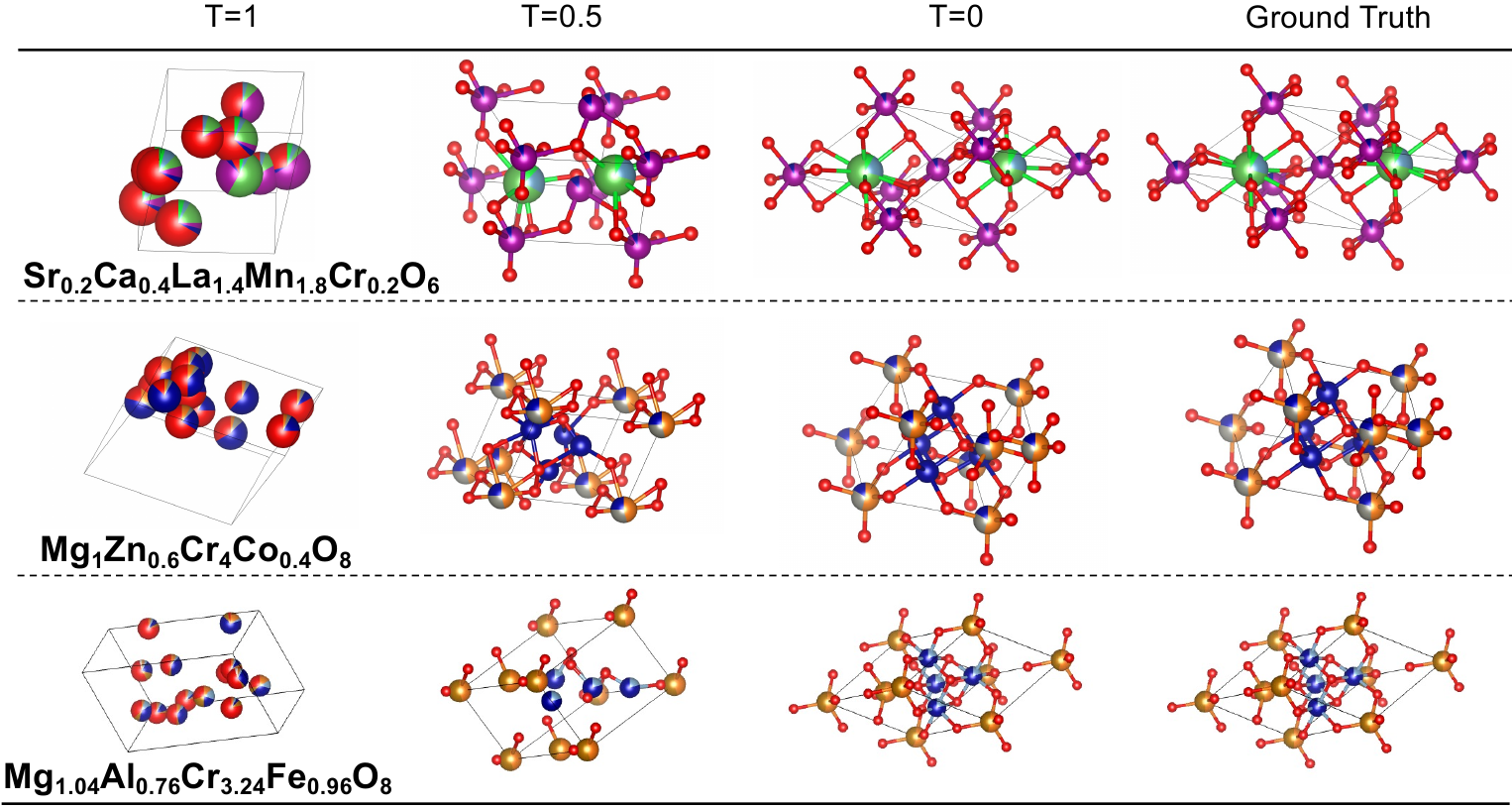}
    \vspace{-0.2cm}
    \caption{
\textbf{Structural reconstruction trajectories of three representative disordered oxides.}
Each row shows the evolution from the noisy state at $T=1$ to the reconstructed structure at $T=0$, compared with the ground truth.
}
    \label{fig:structural_reconstruction}
\end{figure}

Figure~\ref{fig:structural_reconstruction} visualizes representative reconstruction trajectories for three disordered oxide systems.
At $T=1$, the structures are highly noisy and mainly preserve coarse compositional information.
As the trajectory evolves toward $T=0$, EP-Flow progressively recovers ordered atomic positions and site-wise occupancy patterns.
The final predictions are valid, formula-matched, and successfully matched to the corresponding ground-truth structures.

\begin{table}[h]
\centering
\small
\caption{
Reconstruction metrics for the representative disordered oxide cases in Fig.~\ref{fig:structural_reconstruction}.
}
\label{tab:structural_reconstruction_metrics}
\resizebox{\linewidth}{!}{
\begin{tabular}{lccc}
\toprule
\textbf{Formula} & \textbf{RMSD ($\downarrow$)} & \textbf{OccMAE ($\downarrow$)} & \textbf{Max Occ Error ($\downarrow$)} \\
\midrule
$\mathrm{Sr_{0.2}Ca_{0.4}La_{1.4}Mn_{1.8}Cr_{0.2}O_6}$
& 0.0188 & 0.0006 & 0.0019 \\
$\mathrm{Mg_1Zn_{0.6}Cr_4Co_{0.4}O_8}$
& 0.0191 & 0.0053 & 0.0371 \\
$\mathrm{Mg_{1.04}Al_{0.76}Cr_{3.24}Fe_{0.96}O_8}$
& 0.0151 & 0.0052 & 0.0335 \\
\bottomrule
\end{tabular}}
\end{table}

Table~\ref{tab:structural_reconstruction_metrics} summarizes the quantitative reconstruction accuracy.
All three examples achieve RMSD below $0.02$, indicating accurate reconstruction of fractional coordinates and lattice geometry.
The OccMAE values are also small, showing that EP-Flow recovers site-wise fractional occupancies with high fidelity.
The largest occupancy errors mainly occur on mixed cation sites, where multiple chemically similar species compete for the same crystallographic positions.

\begin{table}[h]
\centering
\small
\caption{
Site-wise occupancy comparison for representative structural reconstruction cases.
Only mixed-occupancy sites are reported; fully occupied sites are omitted for clarity.
}
\label{tab:site_occ_comparison}
\resizebox{\linewidth}{!}{
\begin{tabular}{llllc}
\toprule
\textbf{Formula} & \textbf{Mixed site} & \textbf{Reference occupancy} & \textbf{Predicted occupancy} & \textbf{ $\Delta$Occ} \\
\midrule

$\mathrm{Sr_{0.2}Ca_{0.4}La_{1.4}Mn_{1.8}Cr_{0.2}O_6}$
& B-site
& $\mathrm{Mn_{0.900}Cr_{0.100}}$
& $\mathrm{Mn_{0.8987}Cr_{0.1013}}$, $\mathrm{Mn_{0.9013}Cr_{0.0987}}$
& 0.0013 \\

& A-site
& $\mathrm{Sr_{0.100}Ca_{0.200}La_{0.700}}$
& $\mathrm{Sr_{0.1019}Ca_{0.1988}La_{0.6993}}$, $\mathrm{Sr_{0.0981}Ca_{0.2012}La_{0.7007}}$
& 0.0019 \\

\midrule

$\mathrm{Mg_1Zn_{0.6}Cr_4Co_{0.4}O_8}$
& Cation site
& $\mathrm{Mg_{0.500}Zn_{0.300}Co_{0.200}}$
& $\mathrm{Mg_{0.5371}Zn_{0.2726}Co_{0.1903}}$, $\mathrm{Mg_{0.4629}Zn_{0.3274}Co_{0.2097}}$
& 0.0371 \\

\midrule

$\mathrm{Mg_{1.04}Al_{0.76}Cr_{3.24}Fe_{0.96}O_8}$
& Mg/Fe site
& $\mathrm{Mg_{0.520}Fe_{0.480}}$
& $\mathrm{Mg_{0.5535}Fe_{0.4465}}$, $\mathrm{Mg_{0.4865}Fe_{0.5135}}$
& 0.0335 \\

& Al/Cr site
& $\mathrm{Al_{0.190}Cr_{0.810}}$
& $\mathrm{Al_{0.1884\text{--}0.1914}Cr_{0.8086\text{--}0.8116}}$
& 0.0016 \\

\bottomrule
\end{tabular}}
\end{table}

Table~\ref{tab:site_occ_comparison} further compares the reference and predicted occupancies on mixed sites.
For $\mathrm{Sr_{0.2}Ca_{0.4}La_{1.4}Mn_{1.8}Cr_{0.2}O_6}$, EP-Flow almost exactly recovers both A-site Sr/Ca/La disorder and B-site Mn/Cr disorder, with maximum occupancy errors below $0.002$.
This case demonstrates that EP-Flow can accurately reconstruct coupled A/B-site disorder in a perovskite-like oxide framework.

For $\mathrm{Mg_1Zn_{0.6}Cr_4Co_{0.4}O_8}$, the Mg/Zn/Co mixed site is more challenging because three species compete on the same cation sublattice.
The prediction preserves the global formula but slightly redistributes occupancy between the two mixed sites: one site becomes slightly Mg-rich, while the other becomes relatively Zn/Co-rich.
This suggests that the main residual error arises from local redistribution among multiple mixed species rather than failure to satisfy the global composition.

For $\mathrm{Mg_{1.04}Al_{0.76}Cr_{3.24}Fe_{0.96}O_8}$, the Al/Cr mixed sites are reconstructed accurately, with maximum occupancy error around $0.0016$.
The main residual error comes from the Mg/Fe mixed sites, where EP-Flow predicts one site as slightly Mg-rich and the other as slightly Fe-rich.
Importantly, the two predicted Mg/Fe sites still sum to the correct total composition, $\mathrm{Mg_{1.04}Fe_{0.96}}$.
Thus, the error mainly reflects local redistribution among chemically similar mixed sites rather than violation of the global stoichiometry.
Such site-level differentiation is chemically plausible in disordered crystals, although the crystallographic average structure remains the reference for quantitative evaluation.

\subsection{Hyperparameters}\label{app:hyper}

Table~\ref{tab:arch-hparams} summarizes the concrete architecture hyperparameters and optimization-related quantities  exposed by the current implementation.
All experiments were conducted on NVIDIA GeForce RTX 4090 GPUs with 24 GB memory. 
Models on COD were trained on a single GPU, while MPDS models were trained with two-GPU distributed data parallelism due to the larger dataset scale. 
We used 16-bit mixed precision for all training runs.

\begin{table}[h]
\centering
\small
\caption{Architecture and training-related hyperparameters.}
\label{tab:arch-hparams}
\begin{tabular}{@{}p{0.40\linewidth}p{0.20\linewidth}p{0.25\linewidth}@{}}
\toprule
Hyperparameter & Symbol / module & Value \\
\midrule
Hidden dimension & $d_h$ & 512 \\
Message-passing layers & $L$ & 6 \\
Activation function & $\sigma$ & SiLU \\
Learning rate & $lr$ & $5 \times 10^{-4}$ \\
LR decay factor & $\gamma_{\mathrm{lr}}$ & 0.85 \\
LR scheduler patience (epochs) & $P_{\mathrm{lr}}$ & 50  \\
Training epochs & $T_{\mathrm{epoch}}$ & 1000 \\
Sinkhorn iterations & $K_{\mathrm{SK}}$ & 100 \\
ODE steps & $I_{\mathrm{step}}$ & 200 \\
\bottomrule
\end{tabular}
\end{table}

\begin{table}[htbp!]
\centering
\small
\caption{Dataset-specific training hyperparameters and generative loss weights.}
\label{tab:dataset-hparams}
\begin{tabular}{llcccccc}
\toprule
\textbf{Database} & \textbf{Setting} & \textbf{Batch Size} & \textbf{Sampling} $\bm{\mathrm{step\_lr}}$ & $\bm{\lambda_{\mA}}$ & $\bm{\lambda_{\mF}}$ & $\bm{\lambda_{\mathrm{len}}}$ & $\bm{\lambda_{\mathrm{ang}}}$ \\
\midrule
\multirow{2}{*}{COD}
& Substitution & \multirow{2}{*}{256} & \multirow{2}{*}{3} & 20.0 & 100.0 & 1.0 & 100.0 \\
& Disorder     &                      &                    & 20.0 & 100.0 & 1.0 & 100.0 \\
\midrule
\multirow{2}{*}{MPDS}
& Substitution & \multirow{2}{*}{512} & \multirow{2}{*}{5} & 20.0 & 100.0 & 5.0 & 500.0 \\
& Disorder     &                      &                    & 20.0 & 100.0 & 5.0 & 500.0 \\
\bottomrule
\end{tabular}\\[4pt]
\end{table}

\newpage
\section{Limitations}\label{app:limitations}
EP-Flow currently focuses on formula-conditioned generation for disordered crystals.
Although the global stoichiometry provides the main compositional constraint, the generation process also implicitly depends on the number of crystallographic sites.
In practice, the site count is either specified or sampled from the empirical distribution of the benchmark setting.
Therefore, fully determining the site number directly from the formula alone remains an open problem, especially for disordered structures where the summed non-vacancy stoichiometry does not necessarily equal the number of crystallographic sites.

Second, EP-Flow does not currently support property-conditioned generation.
The present framework is designed to generate structurally valid disordered crystals under stoichiometric constraints, but it does not directly optimize target properties such as ionic conductivity, magnetism, superconductivity, or thermodynamic stability.
Extending EP-Flow to property-guided or inverse-design settings is an important direction for future work.

Finally, EP-Flow does not currently condition on experimental measurements such as PXRD, PDF, or spectroscopy data.
Incorporating such information could further constrain the generation process and make the framework more useful for experimental structure determination and refinement.

%% file: reference.bib
@article{merchant2023scaling,
  title={Scaling deep learning for materials discovery},
  author={Merchant, Amil and Batzner, Simon and Schoenholz, Samuel S and Aykol, Muratahan and Cheon, Gowoon and Cubuk, Ekin Dogus},
  journal={Nature},
  volume={624},
  number={7990},
  pages={80--85},
  year={2023},
  publisher={Nature Publishing Group UK London}
}

@article{wu2026dmflow,
  title={DMFlow: Disordered Materials Generation by Flow Matching},
  author={Wu, Liming and Jiao, Rui and Li, Qi and Li, Mingze and Li, Songyou and Jin, Shifeng and Huang, Wenbing},
  journal={arXiv preprint arXiv:2602.04734},
  year={2026}
}

@article{belsky2002icsd,
  title={New developments in the Inorganic Crystal Structure Database (ICSD): accessibility in support of materials research and design},
  author={Belsky, Alec and Hellenbrandt, Mariette and Karen, Vicky Lynn and Luksch, Peter},
  journal={Structural Science},
  volume={58},
  number={3},
  pages={364--369},
  year={2002},
  publisher={International Union of Crystallography}
}

@inproceedings{
lipmanflow,
title={Flow Matching for Generative Modeling},
author={Yaron Lipman and Ricky T. Q. Chen and Heli Ben-Hamu and Maximilian Nickel and Matthew Le},
booktitle={The Eleventh International Conference on Learning Representations },
year={2023},
}

@inproceedings{albergobuilding,
  title={Building Normalizing Flows with Stochastic Interpolants},
  author={Albergo, Michael and Vanden-Eijnden, Eric},
  booktitle={ICLR 2023 Conference},
  year={2023}
}

@article{sinkhorn1964relationship,
  title={A relationship between arbitrary positive matrices and doubly stochastic matrices},
  author={Sinkhorn, Richard},
  journal={The annals of mathematical statistics},
  volume={35},
  number={2},
  pages={876--879},
  year={1964},
  publisher={JSTOR}
}

@inproceedings{
chen2024rfm,
title={Flow Matching on General Geometries},
author={Ricky T. Q. Chen and Yaron Lipman},
booktitle={The Twelfth International Conference on Learning Representations},
year={2024},
}

@article{torgerson1952multidimensional,
  title={Multidimensional scaling: I. Theory and method},
  author={Torgerson, Warren S},
  journal={Psychometrika},
  volume={17},
  number={4},
  pages={401--419},
  year={1952},
  publisher={Springer-Verlag}
}

@article{jiao2023crystal,
  title={Crystal structure prediction by joint equivariant diffusion},
  author={Jiao, Rui and Huang, Wenbing and Lin, Peijia and Han, Jiaqi and Chen, Pin and Lu, Yutong and Liu, Yang},
  journal={Advances in Neural Information Processing Systems},
  volume={36},
  pages={17464--17497},
  year={2023}
}

@inproceedings{miller2024flowmm,
  title={FlowMM: Generating Materials with Riemannian Flow Matching},
  author={Miller, Benjamin Kurt and Chen, Ricky TQ and Sriram, Anuroop and Wood, Brandon M},
  booktitle={International Conference on Machine Learning},
  pages={35664--35686},
  year={2024},
  organization={PMLR}
}

@inproceedings{
xie2022cdvae,
title={Crystal Diffusion Variational Autoencoder for Periodic Material Generation},
author={Tian Xie and Xiang Fu and Octavian-Eugen Ganea and Regina Barzilay and Tommi S. Jaakkola},
booktitle={International Conference on Learning Representations},
year={2022},
}

@article{luo2024cond_cdvae,
  title={Deep learning generative model for crystal structure prediction},
  author={Luo, Xiaoshan and Wang, Zhenyu and Gao, Pengyue and Lv, Jian and Wang, Yanchao and Chen, Changfeng and Ma, Yanming},
  journal={npj Computational Materials},
  volume={10},
  number={1},
  pages={254},
  year={2024},
  publisher={Nature Publishing Group UK London}
}

@article{zeni2025mattergen,
  title={A generative model for inorganic materials design},
  author={Zeni, Claudio and Pinsler, Robert and Z{\"u}gner, Daniel and Fowler, Andrew and Horton, Matthew and Fu, Xiang and Wang, Zilong and Shysheya, Aliaksandra and Crabb{\'e}, Jonathan and Ueda, Shoko and others},
  journal={Nature},
  volume={639},
  number={8055},
  pages={624--632},
  year={2025},
  publisher={Nature Publishing Group UK London}
}

@inproceedings{lin2024equicsp,
  title={Equivariant diffusion for crystal structure prediction},
  author={Lin, Peijia and Chen, Pin and Jiao, Rui and Mo, Qing and Jianhuan, Cen and Huang, Wenbing and Liu, Yang and Huang, Dan and Lu, Yutong},
  booktitle={Forty-first International Conference on Machine Learning},
  year={2024}
}

@article{wu2025dao,
	title = {Siamese foundation models for crystal structure prediction},
	issn = {2041-1723},
	doi = {10.1038/s41467-026-72362-3},
	journal = {Nature Communications},
	author = {Wu, Liming and Huang, Wenbing and Jiao, Rui and Huang, Jianxing and Liu, Liwei and Zhou, Yipeng and Sun, Hao and Liu, Yang and Sun, Fuchun and Ren, Yuxiang and Wen, Ji-Rong},
	year = {2026},
}

@article{luo2024crystalflow,
  title={Crystalflow: A flow-based generative model for crystalline materials},
  author={Luo, Xiaoshan and Wang, Zhenyu and Wang, Qingchang and Lv, Jian and Wang, Lei and Wang, Yanchao and Ma, Yanming},
  journal={arXiv preprint arXiv:2412.11693},
  year={2024}
}

@article{petersen2025dis_gen,
  title={Dis-GEN: Disordered crystal structure generation},
  author={Petersen, Martin Hoffmann and Zhu, Ruiming and Dai, Haiwen and Aggarwal, Savyasanchi and Wei, Nong and Chen, Andy Paul and Bhowmik, Arghya and Lastra, Juan Maria Garcia and Hippalgaonkar, Kedar},
  journal={arXiv preprint arXiv:2507.18275},
  year={2025}
}

@article{kuhn1955hungarian,
  title={The Hungarian method for the assignment problem},
  author={Kuhn, Harold W},
  journal={Naval research logistics quarterly},
  volume={2},
  number={1-2},
  pages={83--97},
  year={1955},
  publisher={Wiley Online Library}
}

@article{wu2025crysbfn,
  title={A periodic bayesian flow for material generation},
  author={Wu, Hanlin and Song, Yuxuan and Gong, Jingjing and Cao, Ziyao and Ouyang, Yawen and Zhang, Jianbing and Zhou, Hao and Ma, Wei-Ying and Liu, Jingjing},
  journal={arXiv preprint arXiv:2502.02016},
  year={2025}
}

@article{ruple2025symmbfn,
  title={Symmetry-Aware Bayesian Flow Networks for Crystal Generation},
  author={Ruple, Laura and Torresi, Luca and Schopmans, Henrik and Friederich, Pascal},
  journal={arXiv preprint arXiv:2502.03146},
  year={2025}
}

@inproceedings{jiao2025mofbfn,
  title={Mof-bfn: Metal-organic frameworks structure prediction via bayesian flow networks},
  author={Jiao, Rui and Wu, Hanlin and Huang, Wenbing and Song, Yuxuan and Ouyang, Yawen and Rong, Yu and Xu, Tingyang and Wang, Pengju and Zhou, Hao and Ma, Wei-Ying and others},
  booktitle={The Thirty-ninth Annual Conference on Neural Information Processing Systems},
  year={2025}
}

@article{joshi2025all,
  title={All-atom diffusion transformers: Unified generative modelling of molecules and materials},
  author={Joshi, Chaitanya K and Fu, Xiang and Liao, Yi-Lun and Gharakhanyan, Vahe and Miller, Benjamin Kurt and Sriram, Anuroop and Ulissi, Zachary W},
  journal={arXiv preprint arXiv:2503.03965},
  year={2025}
}

@inproceedings{yi2026crystaldit,
  title={CrystalDiT: Simple Diffusion Transformers for Crystal Generation},
  author={Yi, Xiaohan and Xu, Guikun and Zhang, Zhong and Liu, Liu and Bian, Yatao and Xiao, Xi and Zhao, Peilin},
  booktitle={Proceedings of the AAAI Conference on Artificial Intelligence},
  volume={40},
  pages={1462--1470},
  year={2026}
}

@article{li2025powder,
  title={Powder diffraction crystal structure determination using generative models},
  author={Li, Qi and Jiao, Rui and Wu, Liming and Zhu, Tiannian and Huang, Wenbing and Jin, Shifeng and Liu, Yang and Weng, Hongming and Chen, Xiaolong},
  journal={Nature Communications},
  volume={16},
  number={1},
  pages={7428},
  year={2025},
  publisher={Nature Publishing Group UK London}
}

@misc{MPDS,
  title={Materials Platform for Data Science},
  author={Blokhin, Evgeny and Villars, Pierre},
  year={2022},
  publisher={MPDS. Accessed on}
}

@book{peyre2019computational,
  title={Computational optimal transport: With applications to data science},
  author={Peyr{\'e}, Gabriel and Cuturi, Marco},
  year={2019},
  publisher={Now Foundations and Trends}
}

@article{cuturi2013sinkhorn,
  title={Sinkhorn distances: Lightspeed computation of optimal transport},
  author={Cuturi, Marco},
  journal={Advances in neural information processing systems},
  volume={26},
  year={2013}
}

@article{de2022riemannian,
  title={Riemannian score-based generative modelling},
  author={De Bortoli, Valentin and Mathieu, Emile and Hutchinson, Michael and Thornton, James and Teh, Yee Whye and Doucet, Arnaud},
  journal={Advances in neural information processing systems},
  volume={35},
  pages={2406--2422},
  year={2022}
}

@article{gat2024discrete,
  title={Discrete flow matching},
  author={Gat, Itai and Remez, Tal and Shaul, Neta and Kreuk, Felix and Chen, Ricky TQ and Synnaeve, Gabriel and Adi, Yossi and Lipman, Yaron},
  journal={Advances in Neural Information Processing Systems},
  volume={37},
  pages={133345--133385},
  year={2024}
}

@article{davis2024fisher,
  title={Fisher flow matching for generative modeling over discrete data},
  author={Davis, Oscar and Kessler, Samuel and Petrache, Mircea and Ceylan, {\.I}smail {\.I} and Bronstein, Michael and Bose, Avishek J},
  journal={Advances in Neural Information Processing Systems},
  volume={37},
  pages={139054--139084},
  year={2024}
}

@inproceedings{stark2024dirichlet,
  title={Dirichlet Flow Matching with Applications to DNA Sequence Design},
  author={Stark, Hannes and Jing, Bowen and Wang, Chenyu and Corso, Gabriele and Berger, Bonnie and Barzilay, Regina and Jaakkola, Tommi},
  booktitle={International Conference on Machine Learning},
  pages={46495--46513},
  year={2024},
  organization={PMLR}
}

@article{fishman2023diffusion,
  title={Diffusion models for constrained domains},
  author={Fishman, Nic and Klarner, Leo and De Bortoli, Valentin and Mathieu, Emile and Hutchinson, Michael},
  journal={arXiv preprint arXiv:2304.05364},
  year={2023}
}

@article{liu2023mirror,
  title={Mirror diffusion models for constrained and watermarked generation},
  author={Liu, Guan-Horng and Chen, Tianrong and Theodorou, Evangelos and Tao, Molei},
  journal={Advances in Neural Information Processing Systems},
  volume={36},
  pages={42898--42917},
  year={2023}
}

@article{utkarsh2025physics,
  title={Physics-constrained flow matching: Sampling generative models with hard constraints},
  author={Utkarsh, Utkarsh and Cai, Pengfei and Edelman, Alan and Gomez-Bombarelli, Rafael and Rackauckas, Christopher Vincent},
  journal={arXiv preprint arXiv:2506.04171},
  year={2025}
}

@article{diederen2025flows,
  title={Flows on convex polytopes},
  author={Diederen, Tomek and Zamboni, Nicola},
  journal={arXiv preprint arXiv:2503.10232},
  year={2025}
}

@article{ong2013pymatgen,
  title={Python Materials Genomics (pymatgen): A robust, open-source python library for materials analysis},
  author={Ong, Shyue Ping and Richards, William Davidson and Jain, Anubhav and Hautier, Geoffroy and Kocher, Michael and Cholia, Shreyas and Gunter, Dan and Chevrier, Vincent L and Persson, Kristin A and Ceder, Gerbrand},
  journal={Computational Materials Science},
  volume={68},
  pages={314--319},
  year={2013},
  publisher={Elsevier}
}

@article{gravzulis2012cod,
  title={Crystallography Open Database (COD): an open-access collection of crystal structures and platform for world-wide collaboration},
  author={Gra{\v{z}}ulis, Saulius and Da{\v{s}}kevi{\v{c}}, Adriana and Merkys, Andrius and Chateigner, Daniel and Lutterotti, Luca and Quiros, Miguel and Serebryanaya, Nadezhda R and Moeck, Peter and Downs, Robert T and Le Bail, Armel},
  journal={Nucleic acids research},
  volume={40},
  number={D1},
  pages={D420--D427},
  year={2012},
  publisher={Oxford University Press}
}

@inproceedings{
hollmer2025omatg,
title={Open Materials Generation with Stochastic Interpolants},
author={Philipp H{\"o}llmer and Thomas Egg and Maya Martirossyan and Eric Fuemmeler and Zeren Shui and Amit Gupta and Pawan Prakash and Adrian Roitberg and Mingjie Liu and George Karypis and Mark Transtrum and Richard Hennig and Ellad B. Tadmor and Stefano Martiniani},
booktitle={Forty-second International Conference on Machine Learning},
year={2025}
}
